\documentclass[11pt]{article}

\usepackage[final]{acl}

\usepackage{times}
\usepackage{latexsym}

\usepackage[T1]{fontenc}

\usepackage[utf8]{inputenc}

\usepackage{microtype}

\usepackage{inconsolata}

\usepackage{graphicx}

\usepackage{amssymb,amsmath,amsfonts}
\usepackage{algorithmic}
\usepackage{algorithm}
\usepackage{graphicx}
\usepackage{array}
\usepackage{textcomp}
\usepackage{stfloats}
\usepackage{url}
\usepackage{verbatim}
\usepackage{color}
\usepackage{multirow}
\usepackage{makecell}
\usepackage{hyperref}
\usepackage{amssymb}
\usepackage{subfigure}
\usepackage{calligra}
\usepackage[normalem]{ulem}
\usepackage[table,xcdraw]{xcolor}
\usepackage{booktabs}
\usepackage{bbding}
\usepackage[shortlabels]{enumitem}
\usepackage{epigraph} 

\title{MissMAC-Bench: Building Solid Benchmark for Missing Modality Issue in Robust Multimodal Affective Computing}

\author{Ronghao Lin$^{1,2}$,
        Honghao Lu$^{2}$, 
        Ruixing Wu$^{2}$,
        Aolin Xiong$^{2}$, 
        Qinggong Chu$^{2}$, \\
        {\bf Qiaolin He}$^{2}$,
        {\bf Sijie Mai}$^{3}$,
        {\bf Haifeng Hu}\thanks{ \quad Corresponding authors.}$^{2,4}$\\
        \textsuperscript{1}College of Computer Science and Software Engineering, Shenzhen University \\
        \textsuperscript{2}School of Electronics and Information Technology, Sun Yat-Sen University \\
        \textsuperscript{3}School of Computer Science, South China Normal University\\
        \textsuperscript{4}Pazhou Laboratory\\
    \texttt{linrh@szu.edu.cn;\ huhaif@mail.sysu.edu.cn}
}

\begin{document}
\maketitle
\begin{abstract}
Current Multimodal Affective Computing (MAC) systems heavily rely on the completeness of multiple modalities to accurately understand human's affective state. However, in real-world scenarios, the availability of modality data is often dynamic and uncertain, leading to substantial performance fluctuations due to the distribution shifts and semantic deficiencies of the incomplete multimodal inputs. Known as the missing modality issue, this challenge poses a critical barrier to the robustness and practical deployment of MAC models. To systematically quantify this issue, we introduce \textbf{MissMAC-Bench}, a comprehensive benchmark designed to establish fair and unified evaluation standards from the perspective of cross-modal synergy. Two guiding principles are proposed, including no missing prior during training, and one single model capable of handling both complete and incomplete modality scenarios, thereby ensuring better generalization. Moreover, to bridge the gap between academic research and real-world applications, our benchmark integrates evaluation protocols with both fixed and random missing patterns at the dataset and instance levels. Extensive experiments conducted on 3 widely-used language models across 4 datasets validate the effectiveness of diverse MAC approaches in tackling the missing modality issue. Our benchmark provides a solid foundation for advancing robust MAC and promotes the development of multimedia data mining. Our code is released in \url{https://github.com/RH-Lin/MissMAC-Bench}.
\end{abstract}

\section{Introduction}

\epigraph{Integration of information from multiple sensory channels is crucial for understanding tendencies and reactions in humans.}{—— Partan and Marler, 1999}

With the rapid growth of online platforms and social media, vast amounts of multimodal data capturing diverse forms of human interaction and communication are now widely available on the web. These data encompass synchronized streams of language, speech, and visual cues, enabling the discovery of affective patterns, behavioral insights, and contextual information embedded within human communication. \cite{baltruvsaitis2018multimodal}. Emerged as a promising research area, Multimodal affective computing (MAC) aims to integrate heterogeneous signals from text, audio and vision modalities in speaking videos and predict human's affective states, such as sentiment scores and emotion classes  \cite{poria2020beneath}. As a subfield of multimedia content analysis, previous MAC models have primarily devoted to learning productive fusion-based multimodal representations by designing delicate multimodal fusion techniques \cite{poria2017review,zhao2024deep}. Given its critical role in real-world applications including digital marketing, distance education, and human-computer interaction \cite{liang2024foundations}, MAC models are expected to exhibit strong generalization and robust performance in downstream inference to enable broader applicability.

\begin{figure}[t]
	\centering 
	\includegraphics[scale=0.4]{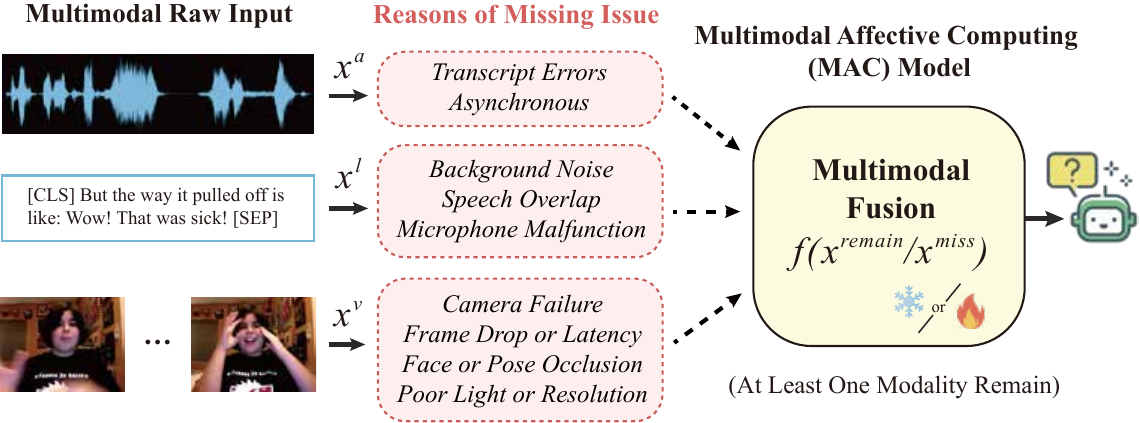}
	\caption{Missing modality issue for robust multimodal affective computing in downstream application.}
	\label{figure_missing_scenario}
\vspace{-0.5cm}
\end{figure}

Most MAC approaches assume that all predefined modalities are present in complete pairs during inference to reach superior performance. However, in downstream applications, the completeness of modalities can not be guaranteed due to numerous reasons including sensor failure or environment interference \cite{wu2024comprehensive,ge2024urrlimvc}, as shown in Figure \ref{figure_missing_scenario}. The lack of modalities leads to significant performance variation and degradation when applying pre-trained models in the real world, named as missing modality issue \cite{ma2022multimodal,liang2024foundations,zhao2025diffmv}. 


Although previous methods have partially mitigated the impact of missing modality\cite{wu2026deep}, there remains a lack of unified evaluation standards. Inconsistent evaluation settings and metrics make it challenging to fairly compare different methods across diverse scenarios \cite{lin2023missmodal,lian2023gcnet}. 
Moreover, prior studies have not clearly distinguished intra- and inter-modal robustness (Details in Appendix \ref{missing_degree_appendix}) during evaluation, limiting the understanding of multimodal model behavior under various missing conditions \cite{wu2024comprehensive}.

In addition, most methods are evaluated only under specific and method-dependent settings, failing to comprehensively cover the holistic spectrum of diverse missing modality scenarios with sufficient flexibility and efficiency \cite{ma2021smil}. Lastly, to enhance the systematical evaluation on model's generalization ability, different MAC sub-tasks and datasets must be carefully considered when addressing the missing modality issue, as MAC requires rich semantic understanding, with diverse affective contributions from each modality \cite{ma2022multimodal,lin2023mtmd}.

To ensure fair comparison, establishing standardized evaluation benchmark is both necessary and critical for advancing research in missing modality issue of MAC. Therefore, focusing on revealing cross-modal synergy, we present a solid benchmark, named \textbf{MissMAC-Bench}, to evaluate models’ inter-modal robustness in MAC task. Our benchmark introduces two main principles to conduct practical robustness assessment and employs both fixed and random missing protocols, with random evaluation conducted at dataset- and instance-level to comprehensively cover diverse missing scenarios. We further provide analysis on 27 current methods to consistently show the applicability ability of different types of methods across multimodal sentiment analysis and emotion recognition subtasks. Our contribution can be summarized as: 

\begin{itemize}[leftmargin=*]
    \item \textbf{Unified evaluation standards}: Two principles are introduced to solidly simulate the performance of pre-trained models in practical applications, including no missing prior and single model to handle various incomplete scenarios;

    \item \textbf{Comprehensive missing modality protocols}: Both fixed and random evaluation protocols are adopted, with random evaluation conducted at both dataset and instance level to bridge the gap between academia and real-world application.

    \item \textbf{Extensive empirical validation}: To facilitate a systematic assessment of generalization capabilities, the benchmark is evaluated across 2 sub-tasks, 4 datasets, and 3 widely-used language models, providing a thorough evaluation of cross-modal synergy and inter-modal robustness.
\end{itemize}

\section{Method Taxonomy}
We categorize previous methods in addressing the missing modality issue into four types of techniques described as follows. 

\subsection{Data Augmentation Methods}
According to task goal, Multimodal affective computing (MAC) can be divided into sentiment score regression and emotion category classification \cite{poria2017review,geetha2024multimodal}. Previous MAC models have primarily devoted to developing diverse multimodal fusion methods, including feature subspace disentanglement \cite{hazarika2020misa}, attention mechanism \cite{rahman2020integrating}, knowledge guidance \cite{yu2021learning,lin2023mtmd}, information theory \cite{han2021improving,xiao2024neuroinspired}, sequence-based recurrent network \cite{majumder2019dialoguernn,hu2021dialoguecrn} and graph neural network \cite{joshi2022cogmen,nguyen2023conversation}. When these models encounter missing modality during downstream inference, their performance drops significantly \cite{hazarika2022analyzing,lin2023missmodal}. A straightforward approach is to apply general data augmentation techniques to enhance robustness while preserve the fusion networks. For instance, \citet{hazarika2022analyzing} introduce dropping and noise into the language modality while \citet{lin2024adapt} apply mixup strategy at both data and feature levels to create augmented data. However, these methods often degrade the original fusion performance, highlighting the need for more effective techniques to balance robustness on incomplete multimodal learning and the pre-trained complete capability.



\subsection{Alignment-based Methods}
Considering input with incomplete and complete modalities as different views of the same multimodal samples, previous methods conduct multi-view learning \cite{zhang2022cpmnet,yu2025merlin} to align the incomplete and complete representations without imputation. Techniques including contrastive learning \cite{poklukar2022geometric,lin2023missmodal}, canonical correlation analysis \cite{andrew2013dcca,wang2015dccae,sun2020learning}, adversarial learning \cite{yuan2023noise} and optimal transport \cite{han2022mmalign} have been explored for this purpose. These alignment-based methods, focus on exploiting the commonly shared and invariant features to compensate missing information. Although they can effectively improve the consistency between complete and incomplete multimodal representations, their performance remains suboptimal due to the huge modality gap in multimodal fusion \cite{liang2022mind}.

\subsection{Generation-based Methods}
Another promising approach is leveraging generative models to synthesize missing modality data, and then using the generated information to supplement the incomplete inputs. Current generation-based methods utilize current generative models include auto-encoders \cite{tran2017cra, pham2019found, zhao2021missing}, variational auto-encoders \cite{wu2018mvae,shi2019mmvae,vasco2022muse,cheng2025scrag}, graph completion networks \cite{lian2023gcnet}, and diffusion models \cite{wang2023incomplete} to translate information across different modalities or generate the missing information. These generative methods effectively reconstruct the missing information due to their powerful imputation capabilities. 
Meanwhile, they can mostly maintain the original multimodal fusion performance when no modifications are required for the original fusion networks. 
However, these methods demand higher computation costs and complex hierarchical networks, making them less practical for real-time applications in real world.

\subsection{Adaption-based Methods}
With the emergence of multimodal large language model \cite{zhang2024mmllm}, leveraging additional parameter-efficient fine-tuning techniques to adapt the original multimodal model into various missing scenarios have also been increasingly explored, such as prompt tuning \cite{lee2023mpmm,guo2024mplmm}, mixture-of-expert networks \cite{xu2024momke,han2024fusemoe,gao2024enhanced}, and adapter module \cite{zhang2024towards,zhu2025proxy}. Besides, since there exists optimization conflict between complete and incomplete multimodal learning and imbalance contribution of diverse modalities, some methods adopt adaptive gradient manipulation \cite{wei2025on,lin2024adapt,zhang2024gradient} to adjust the contribution of each modality in various missing scenarios. Since the adaptation-based methods can adjust existing frameworks with minimal additional parameters or address gradient conflicts, they generally exhibit strong generalization ability on diverse pre-trained models. Thus, an increasing number of recent researches tend to adopt such approaches to enhance the adaptability across various input scenarios.

\section{Task Formulation}
\label{sec_task_formulation}
MissMAC-Bench aims to evaluate the robustness of Multimodal Affective Computing (MAC) models when handling incomplete inputs with missing modality, while still harnessing the potential of multimodal fusion when complete data is available. To simulate the downstream scenarios in the real world, we establish a standardized evaluation setting for both training and inference stages, guided by two main principle: (1) \textbf{No missing prior}; (2) \textbf{One unified model to handle both complete and incomplete input with various missing scenarios}.

We consider the missing modality issue in MAC with three modalities: textual, acoustic, and visual modality, denoted as $u\in\{l,a,v\}$. Given a multimodal data sample $x_i$ as input, the complete observation is defined as $x_M \equiv \{x^l_i,x^a_i,x^v_i\}\in\mathbb{R}^{d_u} $ including language, audio and vision data, respectively, where  where $d_u$ is the dimension of the input vector. When some modalities are missing, the incomplete observation is defined as $x_{m} \equiv \{x^{remain}_i,x^{miss}_i\} $ where $x^{miss}_i$ denotes the missing modality and $x^{remain}_i$ denotes the remained ones. 

During training, both $x^{miss}_i$ and $x^{remain}_i$ are allowed to be utilized for fully exploiting the available multimodal data. Current MAC models have presented sophisticated multimodal fusion modules to integrate information across modalities \cite{poria2017review}, providing an excellent initialization model for handling missing modality. With the pre-trained weights of multimodal fusion modules, the MAC model can be optimized either jointly or separately for both complete and incomplete inputs. 

To claim models' generalizability, the first principle is to deliberately avoid introducing any prior assumptions about the missing modality in the training stage. This is crucial because the missing scenarios in real-world downstream applications are uncertain and can vary widely. Note that most previous methods \cite{lian2023gcnet,wang2023incomplete,zhang2024towards,guo2024mplmm} use the same missing rates or masks to simulate missing scenarios of data in both training and inference stage to maintain higher performance. Nevertheless, we argue that such setting will obtain a prior bias for the missing data to achieve better performance, due to the unknown missing circumstances in the real-world downstream inference \cite{lin2023missmodal}. Besides, simulating missing data during training will cause overfitting on the observed data in each iterations, leading to low training data utilization and unstable loss\cite{hazarika2022analyzing,ma2022multimodal,lin2024adapt}. Therefore, the circumstance of missing modality remains uncertain and unknowable during training in our benchmark.

During inference, the MAC model can operate either in complete multimodal learning with all modalities for samples $x_M$ or in incomplete multimodal learning with one or more missing modalities. The second principle is to train each model only once and use the same pre-trained weights for evaluation across diverse modality-presence scenarios. This setup further simulates real-world scenarios, where the model cannot be modified regardless of the type or completeness of the input data.

For incomplete multimodal learning with missing modality, the $x^{miss}_i$ part is masked out with various missing scenarios during inference. Relying solely on the remained modalities $x^{remain}_i$, the MAC model is expected to   approach the upper bound performance referring to inference with complete modalities.

For complete and incomplete modalities input $x_M$ and $x_m$, the MAC model conduct multimodal fusion $f(\cdot)$ and project the final representations into the affection space to achieve the predicted labels $\hat{y}$, formulated as follows:
\begin{equation}
\hat{y} = \left\{
\begin{aligned}
    & f(x_M)=f(x^l,x^a,x^v) \\
    & f(x_m)=f(x^{remain}/x^{miss})
\end{aligned}
    \right. \in\mathbb{R}^1
\end{equation}

Additionally, the developed MAC model could be deployed for both multimodal sentiment analysis and emotion recognition subtasks, where $\hat{y}$ refers to sentiment score  and emotion class, respectively. As the regression and classification sub-tasks of  MAC, the task optimization objectives of MAC can be computed with the ground truth labels $y_i$ and the affection prediction $\hat{y}_i$:
\begin{equation} \mathcal{L}_{task} \left\{
\begin{aligned}
 &\mathop{=}\limits_{regress} \ \mathop{\mathbb{E}}\limits_{i\in\mathcal{N}}\|y_i-\hat{y}_i\| \text{ or} \mathop{\mathbb{E}}\limits_{i\in\mathcal{N}} \|y_i-\hat{y}_i\|^2_2 \\
 &\mathop{=}\limits_{classify} \ \mathop{\mathbb{E}}\limits_{i\in\mathcal{N}} \ y_i \log \hat{y}_i 
\end{aligned}
    \right.
\end{equation}










\section{Evaluation with Missing Modality Issue}

\begin{figure*}[htbp]
	\centering 
	\includegraphics[scale=0.55]{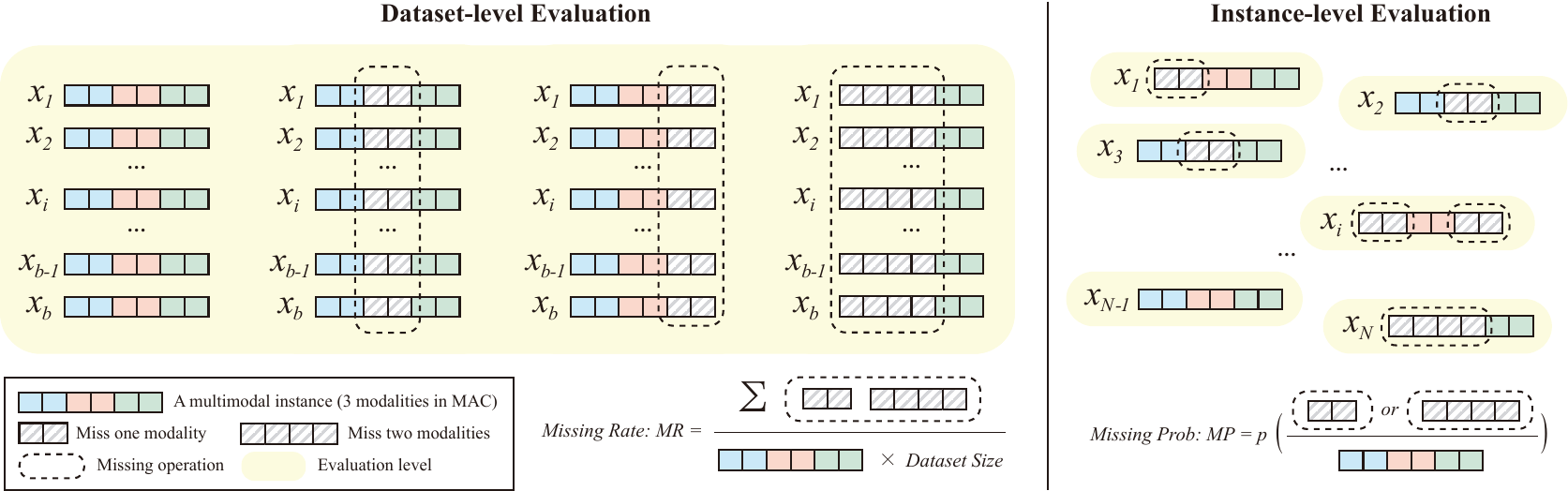}
	\caption{Illustration of random missing protocol, including dataset-level and instance-level evaluation.}
	\label{figure_random_missing_protocol}
\vspace{-0.2cm}
\end{figure*}

Summarized from previous works \cite{zhang2022cpmnet,lin2023missmodal,wang2023incomplete}, we aim to standardize the evaluation settings for the missing modality issue in MAC by setting two protocols.

\subsection{Fixed Missing Protocol}
For the input data in fixed missing protocol, one specific modality (e.g., $\{l,a\}/\{l,v\}/\{a,v\}$) or two modalities (e.g., $\{l\}/\{v\}/\{a\}$) are consistently removed across all iterations, which ensures the same missing patterns being applied either to the entire dataset or to individual instances. The goal of this protocol is to assess the contribution of each modality and bi-modal interactions.

\subsection{Random Missing Protocol}
As shown in Figure \ref{figure_random_missing_protocol}, the random missing protocol evaluates robustness in terms of both efficiency and flexibility through two evaluation levels.

\begin{itemize}[leftmargin=*]
\item \textbf{Dataset-level Evaluation}

We utilize Missing Rate $MR$ to denote the proportion of missing modality relative to the total number of modalities across the entire dataset, formulated as follows: 
\[
MR=\frac{\sum^N_{i=1} k_i}{N\times U} \leq \frac{U-1}{U}
\]
where $N$ denotes total number of samples in datasets, $U=3$ denotes the number of modalities in MAC, $0\leq k_i\leq2$ denotes the number of missing modalities for sample $x_i$. Thus, we set the value of missing rate $MR\in[0.1,0.2,...,0.6,0.7]$ where $0.7$ denotes the most severe scenario with $\frac{2}{3}$ modalities missing.

\item \textbf{Instance-level Evaluation}
    
We adopt Missing Probability $MP$ to represent the probability of each modality being missing within a single multimodal instance. Given one instance sample $x_i$, the probability of missing exactly $k_i$ modalities follows a Binomial Distribution, denoted as: 
\[
p(k_i)=\binom{U}{k_i}MP^{k_i} (1-MP)^{U-k_i}
\]
With $U=3$ and $0\leq k_i\leq2$, we normalize the probabilities to guarantee $p(k_i=0)+p(k_i=1)+p(k_i=2)=1$ and set $MP\in[0.1,0.3,0.5,0.7,0.9]$. Note that $MP=1.0$ denotes the case where each instance has at least one missing modality.

\end{itemize}

To reflect real-world application scenarios, each sample has at least one available modality no matter in which evaluation settings and $MR/MP=0.0$ denotes complete multimodal learning. Due to the space limitation, except for complete modalities input, we only report the performance with $u\in\{l\}/\{a,v\}$ as the dominant and inferior modalities for the fixed missing protocol and with $MR=0.7/MP=1.0$ as the most severe absent situation for the random missing protocol in Section \ref{experiment_section}. More numerical results in various missing circumstances can be found in Appendix \ref{more_exp_appendix}.

\subsection{Evaluation on $C\&R$ Dimension}
Since there are multiple missing circumstances no matter for dataset-level or instance-level evaluation, it's hard and dazzling to compare the performance of different models from one $MR$/$MP$ to another. To better compare the robustness of different methods on random missing protocol as in the real world, we visualize the overall performance and stability of each method as an ellipse as illustrated in Fig. \ref{figure_CR_Dimension} for two dataset on each subtask, along with two complementary evaluation dimensions named as $C\&R$ dimensions:
\begin{figure}[htbp]
	\centering 
	\includegraphics[scale=1.0]{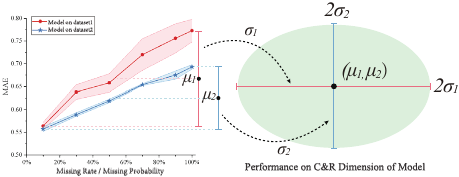}
	\caption{Illustration of $C\&R$ dimensions, where $\mu$ and $\sigma$ denotes the mean and variance of the performance during various missing circumstances. }
	\label{figure_CR_Dimension}
\vspace{-0.5cm}
\end{figure}

\begin{itemize}[leftmargin=*]
\item \textbf{Competence}: This dimension measures the model’s general effectiveness by computing the average performance across all missing modality scenarios in specific missing protocol, providing a holistic view of its ability in handling incomplete inputs. It is denoted as the center of the performance ellipse.

\item \textbf{Resilience}: This dimension captures the performance variation across different missing modality scenarios, ranging from the least to the most severe missing conditions. It is quantified as the standard deviation of performance, reflecting how sensitive a model is to varying degrees of missing modalities. It is denoted as the axis of the performance ellipse.
\end{itemize}

\section{Experiment}\label{experiment_section}
\subsection{Sub-tasks and Datasets}\label{dataset_section}
We employ two public MAC sub-tasks in the proposed benchmark depicted as follows. 

\textbf{Multimodal Sentiment Analysis (MSA):} 
\textit{CMU-MOSI} \cite{zadeh2016mosi} consists of 2,199 monologue utterances from 93 opinion-based YouTube videos by 89 movie reviewers, with sentiment annotations on a continuous scale from $-3$ (strongly negative) to $+3$ (strongly positive). \textit{CMU-MOSEI} \cite{zadeh2018mosei} expands this sentiment datasets by providing about 66 hours video clips from 250 diverse topics and 1,000 unique speakers, with each clip similarly labeled on $[-3, +3]$ Likert scale.

\textbf{Multimodal Emotion Recognition (MER):}
\textit{IEMOCAP} \cite{busso2008iemocap} includes 12 hours of two-way dialogues performed by 10 actors, annotated into six emotions, which are happy, sad, neutral, angry, excited, and frustrated. \textit{MELD} \cite{poria2019meld} comprises approximately 13k utterances from 1,433 multi-party conversations in the TV series Friends, categorized into seven emotion classes including neutral, surprise, fear, sadness, joy, disgust, and angry.

\begin{table*}[htbp]
\caption{Performance comparison on fixed missing protocol with input modalities set as $u\in\{l\}/\{a,v\}$. }
\label{table_exp_fix_l_av}
\centering
\setlength\tabcolsep{4pt}
\scalebox{0.5}{
    \begin{tabular}{c|c|cccccccccccc|cccccccccccc}
    \toprule[1.5pt]
    \multicolumn{2}{c}{Language Model} & \multicolumn{12}{c}{BERT-base / sBERT-base} & \multicolumn{12}{c}{DeBERTaV3-large}\\
    \midrule[1.0pt]
    \multirow{2}{*}{Fix} & \multirow{2}{*}{Models} & \multicolumn{4}{c}{CMU-MOSI} & \multicolumn{4}{c}{CMU-MOSEI} & \multicolumn{2}{c}{IEMOCAP} & \multicolumn{2}{c}{MELD} & \multicolumn{4}{c}{CMU-MOSI} & \multicolumn{4}{c}{CMU-MOSEI} & \multicolumn{2}{c}{IEMOCAP} & \multicolumn{2}{c}{MELD}\\
    \cmidrule(r){3-6}\cmidrule(r){7-10}\cmidrule(r){11-12}\cmidrule(r){13-14}\cmidrule(r){15-18}\cmidrule(r){19-22}\cmidrule(r){23-24}\cmidrule(r){25-26}
    & & Acc7$\uparrow$ & F1$\uparrow$ & MAE$\downarrow$ & Corr$\uparrow$ & Acc7$\uparrow$ & F1$\uparrow$ & MAE$\downarrow$ & Corr$\uparrow$ & Acc$\uparrow$ & wF1$\uparrow$ & Acc$\uparrow$ & wF1$\uparrow$& Acc7$\uparrow$ & F1$\uparrow$ & MAE$\downarrow$ & Corr$\uparrow$ & Acc7$\uparrow$ & F1$\uparrow$ & MAE$\downarrow$ & Corr$\uparrow$ & Acc$\uparrow$ & wF1$\uparrow$ & Acc$\uparrow$ & wF1$\uparrow$\\
    \midrule[1.5pt]
    
     \multirow{27}{*}{$\{l\}$} 
    
     & BBFN$^\tau$ & 44.2 & 84.5 & 0.745 & 0.790 & 51.9 & 85.1 & 0.557 & 0.749 & 51.2 & 52.0 & 59.6 & 58.2  & 
     - & - & - & - & - & - & - & - & - & - & - & - \\
     & MTMD$^\tau$ & 46.7 & 83.5 & 0.764 & 0.782 & 53.1 & 85.2 & 0.544 & 0.753 & 58.0 & 57.5 & 57.7 & 58.0  & 
     - & - & - & - & - & - & - & - & - & - & - & - \\
     & ALMT$^\tau$ & 48.3 & 85.2 & 0.691 & 0.807 & 53.1 & 82.2 & 0.533 & 0.764 & 53.6 & 53.6 & 59.9 & 58.1 & 
     50.7 & 89.5 & 0.576 & 0.883 & 51.6 & 83.0 & 0.551 & 0.777 & 48.8 & 45.5 & 59.2 & 57.2\\
     & ITHP$^\tau$ & 43.8 & 82.6 & 0.762 & 0.784 & 52.1 & 84.7 & 0.551 & 0.756 & 56.5 & 56.2 & 56.3 & 56.3 & 
     49.2 & 87.6 & 0.592 & 0.860 & 50.3 & 83.5 & 0.542 & 0.786 & 50.2 & 49.3 & 57.2 & 56.3\\
     & M3Net$^\tau$ & - & - & - & - & - & - & - & - & 47.3 & 45.5 & 63.4 & 60.4  & 
     - & - & - & - & - & - & - & - & - & - & - & - \\
    
    \cmidrule(r){2-26}
    
    & CCA & 26.4 & 74.7 & 1.098 & 0.567 & 44.6 & 80.2 & 0.677 & 0.621 & 56.1 & 53.9 & 49.7 & 43.1 & 24.5 & 73.9 & 1.145 & 0.552 & 43.6 & 77.1 & 0.727 & 0.536 & 47.6 & 44.5 & 49.2 & 38.9\\
     & DCCA & 23.3 & 73.5 & 1.196 & 0.520 & 41.5 & 68.0 & 0.810 & 0.377 & 51.7 & 50.7 & 48.9 & 35.3 & 15.0 & 32.1 & 1.466 & 0.051 & 41.4 & 56.0 & 0.840 & 0.308 & 27.7 & 18.2 & 48.1 & 31.3\\
     & DCCAE & 25.8 & 66.4 & 1.278 & 0.508 & 41.6 & 67.9 & 0.806 & 0.336 & 50.1 & 49.2 & 47.7 & 32.0 &
     17.9 & 50.2 & 1.358 & 0.179 & 41.2 & 62.8 & 0.804 & 0.304 & 30.7 & 26.5 & 48.1 & 31.3\\
     & CPM-Net & 17.2 & 63.9 & 1.335 & 0.345 & 40.3 & 74.7 & 0.832 & 0.230 & 50.4 & 50.5 & 42.0 & 34.8 &
     15.5 & 40.0 & 1.423 & 0.105 & 41.4 & 48.6 & 0.839 & 0.105 & 23.4 & 14.1 & 35.0 & 30.8\\
     & MM-Align & 36.4 & 82.1 & 0.903 & 0.733 & 46.3 & 82.8 & 0.662 & 0.726 & 52.3 & 49.9 & 40.0 & 36.4  & 
     - & - & - & - & - & - & - & - & - & - & - & - \\
     & NIAT & 38.8 & 82.8 & 0.819 & 0.746 & 54.0 & 85.5 & 0.529 & 0.769 & 62.8 & 62.7 & 59.4 & 60.7  & 
     - & - & - & - & - & - & - & - & - & - & - & - \\
     & MissModal & 44.9 & 84.4 & 0.704 & 0.801 & 52.2 & 85.2 & 0.546 & 0.762 & 62.3 & 62.3 & 57.4 & 57.3 & 
     - & - & - & - & - & - & - & - & - & - & - & - \\

    \cmidrule(r){2-26}
    
     & CRA & 36.3 & 82.8 & 0.900 & 0.749 & 49.3 & 84.9 & 0.576 & 0.741 & 53.8 & 51.3 & 55.3 & 51.6 &
     46.2 & 87.3 & 0.609 & 0.863 & 54.4 & 87.4 & 0.504 & 0.806 & 55.1 & 52.6 & 61.9 & 60.2\\
     & MVAE & 43.0 & 83.8 & 0.731 & 0.786 & 51.7 & 83.8 & 0.565 & 0.761 & 56.3 & 55.8 & 54.6 & 54.8  & 
     - & - & - & - & - & - & - & - & - & - & - & - \\
     & MMVAE & 38.6 & 84.0 & 0.784 & 0.790 & 50.9 & 84.7 & 0.562 & 0.751 & 55.2 & 53.5 & 57.1 & 57.5  & 
     - & - & - & - & - & - & - & - & - & - & - & - \\
     & MCTN & 41.5 & 83.7 & 0.777 & 0.765 & 49.6 & 81.8 & 0.594 & 0.715 & 58.4 & 57.8 & 56.1 & 52.3 &
     48.0 & 87.9 & 0.609 & 0.859 & 53.4 & 86.9 & 0.514 & 0.798 & 60.6 & 59.2 & 34.2 & 29.9\\
     & MMIN & 42.3 & 85.0 & 0.743 & 0.791 & 52.8 & 84.6 & 0.543 & 0.761 & 57.0 & 57.5 & 59.2 & 54.9 &
     45.5 & 88.4 & 0.658 & 0.864 & 53.3 & 86.0 & 0.531 & 0.780 & 56.0 & 55.6 & 16.9 & 4.9\\
     & TFR-Net & 36.7 & 84.5 & 0.801 & 0.773 & 50.9 & 74.7 & 0.583 & 0.744 & 44.4 & 39.1 & 61.7 & 60.7 & 
     44.2 & 86.9 & 0.676 & 0.868 & 46.6 & 85.3 & 0.631 & 0.789 & 28.8 & 20.3 & 47.0 & 36.9\\
     & GCNet & 42.6 & 85.0 & 0.740 & 0.795 & 51.0 & 84.8 & 0.562 & 0.750 & 58.7 & 58.3 & 59.9 & 57.9  & 
     - & - & - & - & - & - & - & - & - & - & - & - \\
     & IMDer & 45.0 & 84.4 & 0.717 & 0.794 & 51.8 & 85.0 & 0.552 & 0.756 & 60.4 & 60.1 & 59.9 & 58.2  & 
     - & - & - & - & - & - & - & - & - & - & - & - \\
    
    \cmidrule(r){2-26}
     & MPLMM & 37.2 & 83.9 & 0.840 & 0.784 & 49.8 & 84.6 & 0.567 & 0.761 & 60.3 & 59.7 & 56.5 & 56.8  & 
     - & - & - & - & - & - & - & - & - & - & - & -\\
     & MoMKE & 43.7 & 80.3 & 0.798 & 0.756 & 52.5 & 85.1 & 0.544 & 0.768 & 60.0 & 59.9 & 56.9 & 56.6  & 
     - & - & - & - & - & - & - & - & - & - & - & -\\
     & FuseMoE & 43.2 & 83.2 & 0.823 & 0.781 & 51.0 & 84.1 & 0.563 & 0.737 & 61.8 & 61.3 & 47.2 & 49.6  & 
     - & - & - & - & - & - & - & - & - & - & - & -\\
     & GMD & 42.3 & 86.0 & 0.770 & 0.800 & 53.1 & 85.1 & 0.538 & 0.765 & 39.5 & 33.1 & 57.6 & 56.5 & 47.8 & 87.4 & 0.614 & 0.862 & 54.7 & 87.4 & 0.508 & 0.799 & 32.9 & 23.7 & 59.6 & 55.1\\
     & EUAR & 46.2 & 85.5 & 0.723 & 0.792 & 52.3 & 84.4 & 0.552 & 0.752 & 55.7 & 55.6 & 58.7 & 57.2 & 
     49.9 & 89.6 & 0.572 & 0.883 & 52.4 & 82.6 & 0.539 & 0.797 & 37.7 & 29.8 & 59.2 & 58.8\\
     & LNLN & 43.9 & 83.5 & 0.761 & 0.760 & 52.5 & 84.3 & 0.547 & 0.765 & 62.8 & 62.4 & 58.3 & 57.1 & 48.0 & 88.4 & 0.601 & 0.877 & 54.1 & 87.2 & 0.510 & 0.807 & 61.4 & 61.3 & 54.8 & 56.3\\
     & P-RMF & 43.0 & 83.3 & 0.749 & 0.788 & 51.3 & 84.7 & 0.555 & 0.766 & 61.5 & 60.6 & 55.8 & 55.4  & 
     - & - & - & - & - & - & - & - & - & - & - & -\\
    
    \midrule[1.5pt]

     \multirow{27}{*}{$\{a,v\}$}
     & BBFN$^\tau$ & 16.0 & 42.3 & 1.385 & -0.005 & 34.0 & 52.6 & 0.896 & 0.452 & 32.6 & 21.1 & 15.8 & 17.3  & 
     - & - & - & - & - & - & - & - & - & - & - & - \\
     & MTMD$^\tau$ & 15.5 & 42.3 & 1.415 & 0.038 & 41.4 & 57.5 & 0.811 & 0.251 & 34.7 & 29.7 & 48.1 & 31.3  & 
     - & - & - & - & - & - & - & - & - & - & - & - \\
     & ALMT$^\tau$ & 15.6 & 42.3 & 1.390 & 0.054 & 19.1 & 48.5 & 1.379 & 0.362 & 30.3 & 21.1 & 44.0 & 34.0 & 
     16.5 & 35.3 & 1.447 & 0.120 & 32.8 & 39.2 & 0.922 & 0.431 & 32.1 & 23.9 & 33.6 & 31.6\\
     & ITHP$^\tau$ & 15.4 & 30.4 & 1.473 & -0.002 & 41.4 & 51.9 & 0.843 & 0.000 & 17.7 & 17.8 & 10.9 & 2.2 &
     18.4 & 33.6 & 1.456 & 0.128 & 33.4 & 41.2 & 0.897 & 0.428 & 31.8 & 24.5 & 34.2 & 32.2\\
     & M3Net$^\tau$ & - & - & - & - & - & - & - & - & 33.7 & 27.1 & 48.8 & 32.5  & 
     - & - & - & - & - & - & - & - & - & - & - & - \\
    
    \cmidrule(r){2-26}
    & CCA & 17.4 & 48.1 & 1.420 & 0.117 & 41.6 & 63.6 & 0.806 & 0.278 & 39.6 & 35.1 & 48.0 & 36.1 & 
    18.7 & 46.9 & 1.445 & 0.102 & 41.4 & 68.4 & 0.798 & 0.322 & 40.9 & 38.1 & 48.1 & 39.2\\
     & DCCA & 17.5 & 53.6 & 1.445 & 0.093 & 41.2 & 73.6 & 0.790 & 0.408 & 35.4 & 33.3 & 46.1 & 33.7 &
     17.4 & 51.2 & 1.374 & 0.172 & 41.1 & 71.2 & 0.792 & 0.359 & 38.6 & 34.5 & 48.1 & 31.3\\
     & DCCAE & 19.4 & 55.2 & 1.569 & 0.046 & 37.9 & 72.7 & 0.796 & 0.404 & 35.5 & 33.2 & 46.5 & 35.7 &
     16.8 & 52.1 & 1.368 & 0.180 & 41.4 & 56.9 & 0.832 & 0.290 & 31.4 & 25.6 & 48.6 & 35.1 \\
     & CPM-Net & 16.8 & 56.2 & 1.367 & 0.330 & 35.2 & 72.0 & 1.395 & 0.041 & 44.5 & 42.4 & 25.6 & 28.9 &
     15.5 & 30.8 & 1.430 & 0.095 & 41.4 & 48.8 & 0.834 & 0.185 & 28.8 & 26.3 & 17.0 & 14.5\\
     & MM-Align & 16.3 & 56.6 & 1.358 & 0.363 & 38.8 & 62.6 & 0.820 & 0.261 & 23.7 & 9.2 & 2.3 & 0.2 & 
     - & - & - & - & - & - & - & - & - & - & - & - \\
     & NIAT & 22.7 & 42.3 & 2.056 & -0.010 & 20.9 & 36.2 & 1.314 & 0.129 & 23.5 & 8.9 & 16.9 & 4.9 &
     17.1 & 34.4 & 1.940 & 0.102 & 24.1 & 56.3 & 1.361 & 0.070 & 10.5 & 2.0 & 48.4 & 31.6\\
     & MissModal & 15.5 & 25.1 & 1.466 & 0.043 & 43.4 & 71.0 & 0.771 & 0.405 & 29.6 & 22.0 & 41.3 & 30.2 & 
     - & - & - & - & - & - & - & - & - & - & - & - \\
    
    \cmidrule(r){2-26}
     & CRA & 18.5 & 55.0 & 1.410 & 0.066 & 40.4 & 73.2 & 0.773 & 0.453 & 44.1 & 42.0 & 45.3 & 40.1 &
     13.9 & 30.4 & 1.679 & -0.051 & 40.9 & 72.5 & 0.766 & 0.436 & 48.0 & 46.0 & 48.5 & 35.3\\
     & MVAE & 17.8 & 38.7 & 1.509 & 0.067 & 41.0 & 49.4 & 0.873 & 0.182 & 33.1 & 33.9 & 45.1 & 34.6 & 
     - & - & - & - & - & - & - & - & - & - & - & - \\
     & MMVAE & 17.3 & 49.6 & 1.416 & 0.103 & 43.0 & 50.5 & 0.813 & 0.367 & 30.4 & 24.8 & 47.6 & 32.4 & 
     - & - & - & - & - & - & - & - & - & - & - & - \\
     & MCTN & 15.5 & 42.2 & 1.431 & 0.018 & 41.4 & 62.9 & 0.842 & 0.102 & 22.7 & 11.5 & 48.8 & 31.6 & 
     15.5 & 25.1 & 1.496 & 0.016 & 41.4 & 48.5 & 0.841 & 0.046 & 23.7 & 9.1 & 12.1 & 2.6 \\
     & MMIN & 20.3 & 52.3 & 1.427 & 0.081 & 39.8 & 71.0 & 0.774 & 0.431 & 48.2 & 47.4 & 48.3 & 31.5 & 
     16.5 & 40.5 & 1.431 & 0.149 & 38.6 & 72.4 & 0.784 & 0.434 & 44.4 & 43.8 & 32.3 & 32.7\\
     & TFR-Net & 18.2 & 58.0 & 1.387 & 0.121 & 39.2 & 67.6 & 0.861 & 0.414 & 41.7 & 40.9 & 22.1 & 24.9 & 
     17.1 & 29.5 & 1.521 & 0.001 & 25.1 & 34.0 & 1.037 & 0.429 & 23.7 & 9.1 & 46.1 & 33.2\\
     & GCNet & 17.9 & 57.2 & 1.363 & 0.385 & 41.2 & 72.2 & 0.805 & 0.393 & 49.0 & 48.1 & 44.7 & 40.9 & 
     - & - & - & - & - & - & - & - & - & - & - & - \\
     & IMDer & 18.4 & 55.1 & 1.419 & 0.386 & 42.0 & 69.1 & 0.788 & 0.376 & 50.0 & 49.9 & 46.7 & 43.0 & 
     - & - & - & - & - & - & - & - & - & - & - & - \\

    \cmidrule(r){2-26}
    & MPLMM & 18.2 & 52.1 & 1.469 & 0.137 & 40.3 & 72.5 & 0.770 & 0.454 & 37.9 & 36.2 & 16.4 & 16.2  & 
     - & - & - & - & - & - & - & - & - & - & - & - \\
     & MoMKE & 19.8 & 54.7 & 1.459 & 0.099 & 42.1 & 72.9 & 0.775 & 0.425 & 48.4 & 47.7 & 13.4 & 11.6  & 
     - & - & - & - & - & - & - & - & - & - & - & - \\
     & FuseMoE & 19.0 & 50.5 & 1.402 & 0.262 & 41.6 & 72.8 & 0.769 & 0.456 & 45.1 & 43.8 & 43.4 & 32.2  & 
     - & - & - & - & - & - & - & - & - & - & - & - \\
     & GMD & 17.6 & 49.1 & 1.414 & 0.098 & 14.8 & 48.5 & 1.439 & 0.427 & 22.0 & 14.2 & 46.1 & 32.6 & 
     21.1 & 42.3 & 1.444 & 0.026 & 12.6 & 20.1 & 1.528 & 0.380 & 24.8 & 11.4 & 35.1 & 32.9\\
     & EUAR & 15.5 & 25.7 & 1.437 & 0.131 & 39.9 & 70.9 & 0.802 & 0.427 & 31.2 & 26.8 & 35.4 & 33.3 & 
     20.6 & 49.8 & 1.503 & 0.075 & 42.4 & 70.2 & 0.785 & 0.385 & 39.9 & 39.7 & 16.9 & 4.9\\
     & LNLN & 15.6 & 48.7 & 1.441 & 0.076 & 40.8 & 71.7 & 0.775 & 0.441 & 51.5 & 50.4 & 35.5 & 31.8 & 
     15.5 & 26.1 & 1.450 & 0.043 & 39.9 & 70.2 & 0.785 & 0.383 & 41.8 & 41.5 & 12.1 & 4.6\\
     & P-RMF & 17.4 & 44.3 & 1.421 & 0.063 & 40.9 & 73.8 & 0.766 & 0.459 & 40.6 & 40.6 & 12.5 & 7.6  & 
     - & - & - & - & - & - & - & - & - & - & - & - \\
    
    \bottomrule[1.5pt]
\end{tabular}
}
\end{table*}
\begin{table*}[htbp]
\caption{Performance comparison on random missing protocol of the input modalities with missing rates $MR$ set as 0.7 in dataset-level evaluation  and $MP$ set as 1.0 in instance-level evaluation. }
\label{table_exp_random_mr7_mp1}
\centering
\setlength\tabcolsep{4pt}
\scalebox{0.5}{
    \begin{tabular}{c|c|cccccccccccc|cccccccccccc}
    \toprule[1.5pt]
    \multicolumn{2}{c}{Language Model} & \multicolumn{12}{c}{BERT-base / sBERT-base} & \multicolumn{12}{c}{DeBERTaV3-large}\\
    \midrule[1.5pt]
    \multirow{2}{*}{Random} & \multirow{2}{*}{Models} & \multicolumn{4}{c}{CMU-MOSI} & \multicolumn{4}{c}{CMU-MOSEI} & \multicolumn{2}{c}{IEMOCAP} & \multicolumn{2}{c}{MELD} & \multicolumn{4}{c}{CMU-MOSI} & \multicolumn{4}{c}{CMU-MOSEI} & \multicolumn{2}{c}{IEMOCAP} & \multicolumn{2}{c}{MELD}\\
    \cmidrule(r){3-6}\cmidrule(r){7-10}\cmidrule(r){11-12}\cmidrule(r){13-14}\cmidrule(r){15-18}\cmidrule(r){19-22}\cmidrule(r){23-24}\cmidrule(r){25-26}
    & & Acc7$\uparrow$ & F1$\uparrow$ & MAE$\downarrow$ & Corr$\uparrow$ & Acc7$\uparrow$ & F1$\uparrow$ & MAE$\downarrow$ & Corr$\uparrow$ & Acc$\uparrow$ & wF1$\uparrow$ & Acc$\uparrow$ & wF1$\uparrow$& Acc7$\uparrow$ & F1$\uparrow$ & MAE$\downarrow$ & Corr$\uparrow$ & Acc7$\uparrow$ & F1$\uparrow$ & MAE$\downarrow$ & Corr$\uparrow$ & Acc$\uparrow$ & wF1$\uparrow$ & Acc$\uparrow$ & wF1$\uparrow$\\

    \midrule[1.5pt]
     \multirow{27}{*}{MR=0.7} 

     & BBFN$^\tau$ & 25.3 & 61.3 & 1.189 & 0.430 & 41.4 & 56.9 & 0.794 & 0.492 & 36.3 & 33.1 & 28.2 & 32.4 & 
     - & - & - & - & - & - & - & - & - & - & - & - \\
     & MTMD$^\tau$ & 26.3 & 62.2 & 1.191 & 0.459 & 45.2 & 67.5 & 0.735 & 0.447 & 38.4 & 37.7 & 51.2 & 43.7 & 
     - & - & - & - & - & - & - & - & - & - & - & -  \\
     & ALMT$^\tau$ & 25.9 & 61.9 & 1.169 & 0.462 & 32.5 & 65.1 & 1.098 & 0.309 & 35.5 & 32.0 & 50.7 & 44.2 & 
     27.4 & 55.7 & 1.176 & 0.499 & 34.4 & 49.2 & 0.891 & 0.469 & 34.7 & 29.7 & 48.2 & 43.5\\
     & ITHP$^\tau$ & 24.9 & 53.4 & 1.229 & 0.441 & 45.0 & 64.8 & 0.748 & 0.425 & 30.3 & 30.5 & 25.9 & 27.5 &
     26.3 & 54.6 & 1.210 & 0.452 & 33.9 & 48.9 & 0.856 & 0.503 & 25.4 & 24.8 & 43.9 & 43.3\\
     & M3Net$^\tau$ & - & - & - & - & - & - & - & - & 36.5 & 35.1 & 50.3 & 45.4 & 
     - & - & - & - & - & - & - & - & - & - & - & - \\
     
    \cmidrule(r){2-26}
    & CCA & 19.9 & 59.7 & 1.298 & 0.326 & 42.7 & 68.3 & 0.775 & 0.409 & 41.2 & 39.8 & 48.5 & 37.7 &
    18.9 & 55.5 & 1.340 & 0.269 & 42.1 & 69.5 & 0.785 & 0.377 & 38.8 & 34.8 & 48.4 & 37.0\\
     & DCCA & 19.6 & 58.6 & 1.343 & 0.214 & 41.3 & 61.9 & 0.806 & 0.360 & 36.2 & 35.0 & 47.2 & 32.5 &
     17.0 & 45.4 & 1.478 & 0.070 & 40.5 & 63.8 & 0.828 & 0.232 & 28.3 & 24.4 & 48.1 & 31.3\\
     & DCCAE & 18.6 & 55.5 & 1.539 & 0.209 & 38.8 & 55.9 & 0.833 & 0.277 & 35.7 & 32.7 & 47.8 & 33.2 &
     15.9 & 36.5 & 1.439 & 0.123 & 41.1 & 55.1 & 0.829 & 0.224 & 26.3 & 23.1 & 48.2 & 32.1\\
     & CPM-Net & 17.1 & 63.8 & 1.361 & 0.291 & 34.2 & 73.9 & 1.692 & 0.104 & 52.2 & 52.5 & 41.6 & 34.1 &
     15.5 & 26.1 & 1.440 & 0.087 & 41.2 & 49.8 & 0.847 & 0.075 & 27.0 & 20.4 & 32.6 & 31.4\\
     & MM-Align & 24.4 & 60.0 & 1.274 & 0.398 & 37.4 & 68.8 & 0.796 & 0.420 & 32.5 & 32.3 & 20.2 & 23.8  & 
     - & - & - & - & - & - & - & - & - & - & - & - \\
     & NIAT & 27.8 & 59.8 & 1.651 & 0.242 & 23.8 & 50.6 & 1.363 & 0.252 & 36.7 & 34.0 & 31.2 & 30.7 &
     28.0 & 55.4 & 1.299 & 0.435 & 43.2 & 46.8 & 0.856 & 0.306 & 27.9 & 30.5 & 51.8 & 44.9\\
     & MissModal & 25.8 & 52.8 & 1.206 & 0.452 & 45.8 & 70.6 & 0.714 & 0.494 & 37.1 & 36.2 & 32.1 & 37.0 & 
     - & - & - & - & - & - & - & - & - & - & - & - \\
    
    \cmidrule(r){2-26}
    & CRA & 23.6 & 57.3 & 1.311 & 0.401 & 43.0 & 76.5 & 0.713 & 0.548 & 40.1 & 40.0 & 48.7 & 42.7 & 
    27.3 & 61.0 & 1.247 & 0.372 & 44.7 & 76.5 & 0.7 & 0.569 & 45.0 & 42.8 & 52.5 & 44.9\\
     & MVAE & 25.4 & 57.4 & 1.253 & 0.410 & 43.8 & 59.4 & 0.791 & 0.424 & 35.9 & 35.9 & 46.1 & 41.6  & 
     - & - & - & - & - & - & - & - & - & - & - & - \\
     & MMVAE & 24.2 & 59.7 & 1.218 & 0.452 & 44.9 & 64.7 & 0.744 & 0.454 & 35.5 & 33.3 & 50.9 & 43.2  & 
     - & - & - & - & - & - & - & - & - & - & - & - \\
     & MCTN & 23.5 & 56.1 & 1.223 & 0.432 & 43.7 & 64.9 & 0.760 & 0.413 & 28.8 & 30.8 & 51.2 & 41.2 &
     26.8 & 54.7 & 1.179 & 0.480 & 45.6 & 66.2 & 0.728 & 0.466 & 31.1 & 31.6 & 21.1 & 22.0\\
     & MMIN & 25.6 & 59.0 & 1.212 & 0.440 & 43.8 & 75.3 & 0.707 & 0.545 & 45.9 & 45.8 & 51.7 & 41.9 &
     26.5 & 60.3 & 1.179 & 0.470 & 43.7 & 76.0 & 0.704 & 0.558 & 40.8 & 40.9 & 23.0 & 20.3\\
     & TFR-Net & 22.8 & 64.1 & 1.207 & 0.400 & 43.4 & 62.3 & 0.782 & 0.406 & 37.9 & 34.6 & 34.8 & 39.6 &
     25.2 & 61.5 & 1.234 & 0.367 & 37.2 & 52.3 & 0.877 & 0.418 & 25.2 & 14.4 & 47.2 & 34.9\\
     & GCNet & 26.3 & 63.9 & 1.166 & 0.494 & 43.3 & 74.5 & 0.733 & 0.542 & 47.3 & 47.2 & 39.9 & 41.8  & 
     - & - & - & - & - & - & - & - & - & - & - & - \\
     & IMDer & 25.0 & 58.1 & 1.226 & 0.426 & 44.8 & 69.1 & 0.732 & 0.470 & 47.7 & 48.2 & 46.3 & 43.0 & 
     - & - & - & - & - & - & - & - & - & - & - & - \\

    \cmidrule(r){2-26}
     & MPLMM & 24.7 & 64.5 & 1.251 & 0.433 & 42.7 & 75.0 & 0.712 & 0.554 & 39.3 & 39.4 & 27.9 & 32.3 & 
     - & - & - & - & - & - & - & - & - & - & - & -\\
     & MoMKE & 26.0 & 62.4 & 1.212 & 0.411 & 45.8 & 75.4 & 0.700 & 0.540 & 47.1 & 46.9 & 42.8 & 42.0 & 
     - & - & - & - & - & - & - & - & - & - & - & -\\
     & FuseMoE & 30.0 & 55.6 & 1.436 & 0.425 & 45.4 & 74.7 & 0.706 & 0.533 & 44.8 & 44.4 & 44.7 & 40.7 & 
     - & - & - & - & - & - & - & - & - & - & - & -\\
     & GMD & 24.7 & 58.9 & 1.203 & 0.468 & 29.5 & 65.9 & 1.123 & 0.295 & 24.6 & 19.7 & 51.1 & 43.4 & 
     29.9 & 61.6 & 1.177 & 0.433 & 26.3 & 50.5 & 1.203 & 0.305 & 26.9 & 15.8 & 46.1 & 41.0 \\
     & EUAR & 25.6 & 51.7 & 1.207 & 0.446 & 45.3 & 72.2 & 0.717 & 0.503 & 37.9 & 36.5 & 46.6 & 42.7  &
     27.9 & 60.3 & 1.288 & 0.411 & 34.1 & 61.0 & 0.925 & 0.354 & 32.8 & 29.7 & 23.5 & 29.8 \\
     & LNLN & 26.1 & 62.7 & 1.200 & 0.472 & 45.0 & 74.8 & 0.705 & 0.536 & 46.9 & 45.9 & 48.5 & 43.1 &
     25.8 & 53.2 & 1.175 & 0.491 & 45.2 & 71.7 & 0.711 & 0.513 & 41.2 & 41.5 & 26.5 & 28.2 \\
     & P-RMF & 25.3 & 55.4 & 1.209 & 0.450 & 44.6 & 75.6 & 0.703 & 0.547 & 41.6 & 41.4 & 26.5 & 28.5 & 
     - & - & - & - & - & - & - & - & - & - & - & -\\
    
    \midrule[1.0pt]
     \multirow{27}{*}{MP=1.0} 
    
     & BBFN$^\tau$ & 30.4 & 68.8 & 1.071 & 0.548 & 43.9 & 66.9 & 0.726 & 0.574 & 42.1 & 41.0 & 36.6 & 40.9  & 
     - & - & - & - & - & - & - & - & - & - & - & -\\
     & MTMD$^\tau$ & 31.8 & 69.5 & 1.071 & 0.569 & 47.4 & 72.9 & 0.682 & 0.547 & 45.1 & 45.0 & 52.9 & 48.4  & 
     - & - & - & - & - & - & - & - & - & - & - & -\\
     & ALMT$^\tau$ & 31.7 & 68.7 & 1.050 & 0.563 & 37.1 & 70.5 & 0.959 & 0.399 & 42.5 & 41.5 & 52.2 & 48.0 & 
     33.8 & 66.8 & 1.000 & 0.628 & 40.8 & 61.1 & 0.771 & 0.565 & 42.8 & 40.4 & 49.0 & 47.0 \\
     & ITHP$^\tau$ & 30.6 & 62.2 & 1.099 & 0.556 & 46.9 & 70.8 & 0.696 & 0.532 & 37.1 & 37.2 & 34.0 & 37.0 &
     33.1 & 65.2 & 1.024 & 0.605 & 40.2 & 60.8 & 0.752 & 0.581 & 36.7 & 35.5 & 45.9 & 43.1 \\
     & M3Net$^\tau$ & - & - & - & - & - & - & - & - & 45.7 & 45.6 & 53.5 & 49.6  & 
     - & - & - & - & - & - & - & - & - & - & - & -\\
     
    \cmidrule(r){2-26}
    & CCA & 22.2 & 65.6 & 1.239 & 0.403 & 43.5 & 72.9 & 0.745 & 0.490 & 46.9 & 46.2 & 48.8 & 40.1 &
    20.0 & 60.9 & 1.296 & 0.342 & 42.7 & 72.2 & 0.761 & 0.439 & 44.0 & 41.3 & 48.7 & 39.5 \\
     & DCCA & 21.8 & 61.2 & 1.323 & 0.296 & 41.4 & 68.1 & 0.799 & 0.371 & 42.0 & 41.3 & 47.0 & 33.1 &
     17.2 & 46.8 & 1.470 & 0.088 & 40.5 & 65.9 & 0.817 & 0.280 & 32.1 & 29.0 & 48.1 & 31.3\\
     & DCCAE & 20.3 & 60.4 & 1.541 & 0.241 & 38.8 & 64.0 & 0.817 & 0.323 & 41.0 & 39.6 & 47.2 & 33.7 & 
     16.6 & 40.5 & 1.429 & 0.156 & 41.4 & 58.8 & 0.815 & 0.268 & 28.4 & 26.4 & 48.2 & 32.5 \\
     & CPM-Net & 16.9 & 64.1 & 1.335 & 0.339 & 34.0 & 73.6 & 1.400 & 0.066 & 52.6 & 53.1 & 41.6 & 34.4 &  
     15.5 & 25.7 & 1.444 & 0.076 & 41.0 & 50.5 & 1.255 & 0.018 & 26.7 & 19.4 & 46.2 & 34.1\\
     & MM-Align & 20.2 & 61.3 & 1.418 & 0.401 & 39.0 & 72.0 & 0.789 & 0.494 & 35.6 & 35.4 & 23.3 & 26.8 & 
     - & - & - & - & - & - & - & - & - & - & - & -\\
     & NIAT & 30.9 & 67.4 & 1.426 & 0.347 & 31.0 & 60.7 & 1.161 & 0.334 & 42.7 & 42.2 & 38.0 & 39.1 & 
     34.0 & 66.0 & 1.111 & 0.554 & 44.2 & 55.6 & 0.813 & 0.378 & 36.3 & 40.0 & 53.7 & 49.5\\
     & MissModal & 31.8 & 63.4 & 1.060 & 0.569 & 47.8 & 76.0 & 0.662 & 0.589 & 44.4 & 43.7 & 39.6 & 43.4 & 
     - & - & - & - & - & - & - & - & - & - & - & -\\
    
    \cmidrule(r){2-26}
     & CRA & 26.9 & 64.7 & 1.122 & 0.505 & 45.3 & 78.6 & 0.671 & 0.613 & 47.2 & 46.8 & 50.7 & 46.2 & 
     31.3 & 66.5 & 1.107 & 0.504 & 47.5 & 79.7 & 0.639 & 0.640 & 50.4 & 48.3 & 55.3 & 49.9\\
     & MVAE & 31.0 & 66.1 & 1.102 & 0.531 & 46.1 & 66.5 & 0.727 & 0.541 & 42.7 & 43.1 & 48.9 & 45.8 & 
     - & - & - & - & - & - & - & - & - & - & - & -\\
     & MMVAE & 28.3 & 67.8 & 1.095 & 0.567 & 46.9 & 71.3 & 0.691 & 0.556 & 42.6 & 41.1 & 52.4 & 47.8 & 
     - & - & - & - & - & - & - & - & - & - & - & -\\
     & MCTN & 28.1 & 63.4 & 1.115 & 0.535 & 44.7 & 70.3 & 0.719 & 0.505 & 36.4 & 38.7 & 52.3 & 44.3 & 
     32.3 & 64.8 & 1.031 & 0.594 & 47.3 & 72.4 & 0.678 & 0.565 & 38.5 & 38.2 & 25.8 & 25.6\\
     & MMIN & 30.6 & 67.3 & 1.092 & 0.537 & 46.2 & 77.9 & 0.660 & 0.614 & 48.4 & 46.8 & 54.3 & 46.4 &
     31.0 & 67.9 & 1.041 & 0.598 & 46.1 & 79.1 & 0.659 & 0.631 & 47.3 & 47.4 & 25.0 & 24.6\\
     & TFR-Net & 28.1 & 69.4 & 1.096 & 0.510 & 46.0 & 70.1 & 0.719 & 0.517 & 44.8 & 44.0 & 41.4 & 46.1 & 
     29.6 & 67.3 & 1.094 & 0.528 & 40.1 & 62.5 & 0.794 & 0.512 & 31.9 & 26.1 & 46.0 & 35.2\\
     & GCNet & 27.6 & 68.7 & 1.089 & 0.549 & 43.8 & 78.8 & 0.742 & 0.546 & 54.0 & 54.1 & 44.1 & 45.3 & 
     - & - & - & - & - & - & - & - & - & - & - & -\\
     & IMDer & 29.5 & 65.8 & 1.084 & 0.541 & 46.6 & 74.4 & 0.689 & 0.568 & 55.4 & 54.9 & 49.7 & 46.5 & 
     - & - & - & - & - & - & - & - & - & - & - & -\\

    \cmidrule(r){2-26}
    & MPLMM & 28.3 & 70.6 & 1.127 & 0.560 & 44.9 & 77.9 & 0.672 & 0.618 & 46.9 & 46.9 & 36.3 & 40.6 & 
     - & - & - & - & - & - & - & - & - & - & - & -\\
     & MoMKE & 30.7 & 67.1 & 1.118 & 0.495 & 47.3 & 78.6 & 0.653 & 0.612 & 52.8 & 52.7 & 42.2 & 43.7 & 
     - & - & - & - & - & - & - & - & - & - & - & -\\
     & FuseMoE & 32.4 & 66.1 & 1.152 & 0.576 & 46.6 & 77.2 & 0.669 & 0.596 & 50.9 & 50.5 & 45.5 & 43.9 & 
     - & - & - & - & - & - & - & - & - & - & - & -\\
     & GMD & 29.4 & 68.0 & 1.093 & 0.561 & 35.7 & 71.7 & 0.976 & 0.387 & 32.6 & 30.3 & 52.6 & 47.5 &
     35.4 & 69.8 & 1.033 & 0.549 & 33.7 & 61.8 & 1.020 & 0.408 & 35.4 & 31.7 & 47.7 & 44.6\\
     & EUAR & 31.6 & 61.7 & 1.083 & 0.548 & 47.6 & 76.3 & 0.665 & 0.594 & 42.9 & 42.5 & 49.5 & 47.3 &
     34.5 & 69.1 & 1.089 & 0.542 & 41.8 & 71.0 & 0.781 & 0.512 & 40.5 & 39.9 & 30.8 & 39.4\\
     & LNLN & 29.8 & 66.9 & 1.102 & 0.531 & 46.9 & 77.9 & 0.662 & 0.612 & 54.4 & 54.9 & 49.2 & 46.3 & 
     31.3 & 63.8 & 1.027 & 0.612 & 47.2 & 77.0 & 0.656 & 0.612 & 48.5 & 48.9 & 33.4 & 36.6\\
     & P-RMF & 30.4 & 65.2 & 1.075 & 0.563 & 46.3 & 78.4 & 0.662 & 0.617 & 48.6 & 48.5 & 34.3 & 37.2& 
     - & - & - & - & - & - & - & - & - & - & - & -\\

    \bottomrule[1.5pt]
    
\end{tabular}
}
\end{table*}

\subsection{Evaluation Metrics}
For MSA \cite{poria2017review}, the evaluation metrics include seven-class accuracy/non-zero F1-score (\textit{Acc7/F1}) and mean absolute error/Pearson correlation (\textit{MAE/Corr}). \textit{Acc7} measures the average accuracy of sentiment predictions within the label range $[-3, -2.5, -1.5, -0.5, +0.5, +1.5, +2.5, +3]$ loosing the constraints of neutral labels and high intensity, while \textit{F1} evaluates binary positive-negative polarity classification without neutral sentiment. \textit{MAE} quantifies the mean absolute deviation between predictions and ground truth labels, and \textit{Corr} assesses the variation degree of predictions relative to the ground truth.

For MER \cite{geetha2024multimodal}, binary accuracy (\textit{Acc}) and weighted F1-score (\textit{wF1}) are utilized to evaluate the classification performance for overall classes, accounting for the relative quantity of each emotion class to considering label imbalance. 

\subsection{Comparison Models}
We categorize the comparison models as follows:
\begin{itemize}[leftmargin=*]
    \item \textbf{Data augmentation based Models:}
    The state-of-the-art MAC models under robust training with random masking data \cite{hazarika2022analyzing}, denoted as $^\tau$, including BBFN \cite{han2021bi}, ALMT \cite{zhang2023almt}, MTMD \cite{lin2023mtmd}, ITHP \cite{xiao2024neuroinspired} for MSA and COGMEN \cite{joshi2022cogmen}, CORECT \cite{nguyen2023conversation}, M3Net \cite{chen2023m3net} for MER;

    \item \textbf{Alignment-based Models:} CCA \cite{hotelling1992cca},
    DCCA \cite{andrew2013dcca}, DCCAE \cite{wang2015dccae}, MM-Align \cite{han2022mmalign}, CPM-Net \cite{zhang2022cpmnet}, 
    NIAT \cite{yuan2023noise},
    MissModal \cite{lin2023missmodal};
    
    \item \textbf{Generation-based Models:} CRA\cite{tran2017cra}, MVAE \cite{wu2018mvae}, MMVAE \cite{shi2019mmvae}, MCTN \cite{pham2019found}, MMIN \cite{zhao2021missing}, TFR-Net \cite{yuan2021tfrnet}, GCNet \cite{lian2023gcnet}, IMDer \cite{wang2023incomplete};

    \item \textbf{Adaption-based Models:}  MPLMM \cite{guo2024mplmm}, MoMKE \cite{xu2024momke}, FuseMoE \cite{han2024fusemoe}, GMD \cite{zhang2024gradient}, EUAR \cite{gao2024enhanced}, LNLN \cite{zhang2024towards}, P-RMF \cite{zhu2025proxy}.


    





\end{itemize}

Note that except for CPM-Net \cite{zhang2022cpmnet} and GCNet \cite{lian2023gcnet} demanding of missing prior in training, all experiments are conducted strictly satisfying the two main principle presented in Section \ref{sec_task_formulation}. Since MAC highly rely on the performance of text modality due to the rich semantics \cite{gkoumas2021makes}, we reproduce all above models with acoustic and visual features extracted by ImageBind \cite{girdhar2023imagebind}, and textual feature extracted by three widely-used language models with various parameters, including BERT-base (110M) \cite{devlin2019bert} for MSA, sBERT-base (83M) \cite{reimers2019sbert} for MER and DeBERTaV3-large (434M) \cite{he2021deberta,he2023debertav3} for both subtasks. Note that DeBERTaV3-large results are only reported for representative methods that could be feasibly adapted to this stronger encoder. 
We conduct evaluation on 10 random seeds for all models to denote sufficient missing circumstances and report the average performance as the stable results.



\subsection{Performance on Fixed Missing Protocol:} We report experiment results of each model on fixed missing protocol with only textual modality $u\in\{l\}$ and both acoustic and visual modalities $u\in\{a,v\}$ to reveal the contribution of each modality and denote the model behavior on learning unimodal features, as shown in Table \ref{table_exp_fix_l_av}.

\textbf{RQ1: What role do different modalities play? } Textual modality dominates performance significantly while acoustic/vision modalities remain inferior for all models. When only language is available, models achieve substantially higher performance compared to audio and video only conditions, demonstrating that textual information carries the most discriminative affective signals due to its richer semantic information.

\textbf{RQ2: How does the scale of language model affect the modality imbalance?} Larger language models may amplify modality imbalance rather than mitigate it. DeBERTaV3 shows even greater performance disparity between language-present and language-absent conditions compared to BERT/sBERT. This observation suggests that more powerful language models increase reliance on textual features while contributing less on other modalities, thereby exacerbating the modality imbalance issue in the incomplete scenarios.

\textbf{RQ3: What type of methods perform more balanced when limited modalities are presented? } Alignment-based methods demonstrate superior robustness when limited modalities are presented including NIAT and MissModal. Besides, CPM-Net maintain relatively balanced performance across different fixed missing patterns. While adaption-based methods perform least effectively in language-absent scenarios with severe performance drops when text is missing. This pattern occurs because alignment-based methods explicitly learn to transfer cross-modal synergy information among different modalities, benefiting both MSA and MER subtasks regardless of which modality is available, while adaptation-based methods with limited parameters tend to overfit on the dominant textual modality, achieving strong performance when language is present but degrading significantly when it is absent.

\subsection{Performance on Random Missing Protocol:} We report experiment results on random missing protocol with the most severe missing scenarios MR=0.7 and MP=1.0, as shown in Table \ref{table_exp_fix_l_av}. 

\textbf{RQ1: How does the evaluation level affect the performance? }
In the most severe missing scenario, instance-level evaluation (MP=1.0) shows consistently better performance than dataset-level evaluation (MR=0.7) across all methods. This discrepancy arises because MR=0.7 simulates missing issue by removing the most modalities in the whole dataset, which causes huger performance degradation across all models with least information provided in each sample. Conversely, instance-level evaluation better matches real-world scenarios where input queries are presented instance by instance with dynamic missing circumstance, making it more practical and less performance drop with MP=1.0 as the most severe scenario.

\textbf{RQ2: How does larger language model impact the results with random missing scenarios? }  Larger language models do not consistently improve performance in incomplete multimodal learning due to greater modality imbalance. Although textual modality contributes significantly in MAC task, the drawback of imbalanced contribution among different modalities may largely cause fluctuations when there are incomplete modality inputs, as stronger language encoders underfitting information of inferior modalities and providing less learning on the cross-modal synergy.

\textbf{RQ3: What type of methods perform better in random missing scenarios? } Adaptation-based and generation-based methods contribute most effectively in random missing scenarios benefiting from either adaptive mechanisms dynamically adjusting to varying missing patterns or directly reconstruct missing information as imputation for multimodal fusion. While alignment-based methods can partially enhance incomplete robustness, they struggle with suboptimal performance in the severest missing scenarios due to their reliability on the static fusion representation.

\begin{figure*}[htbp]
\centering
\subfigure[with $MR$ on MSA]{
\label{fig_vis_CR_dimension_BERT_MSA_MR}
\includegraphics[width=0.23\textwidth]{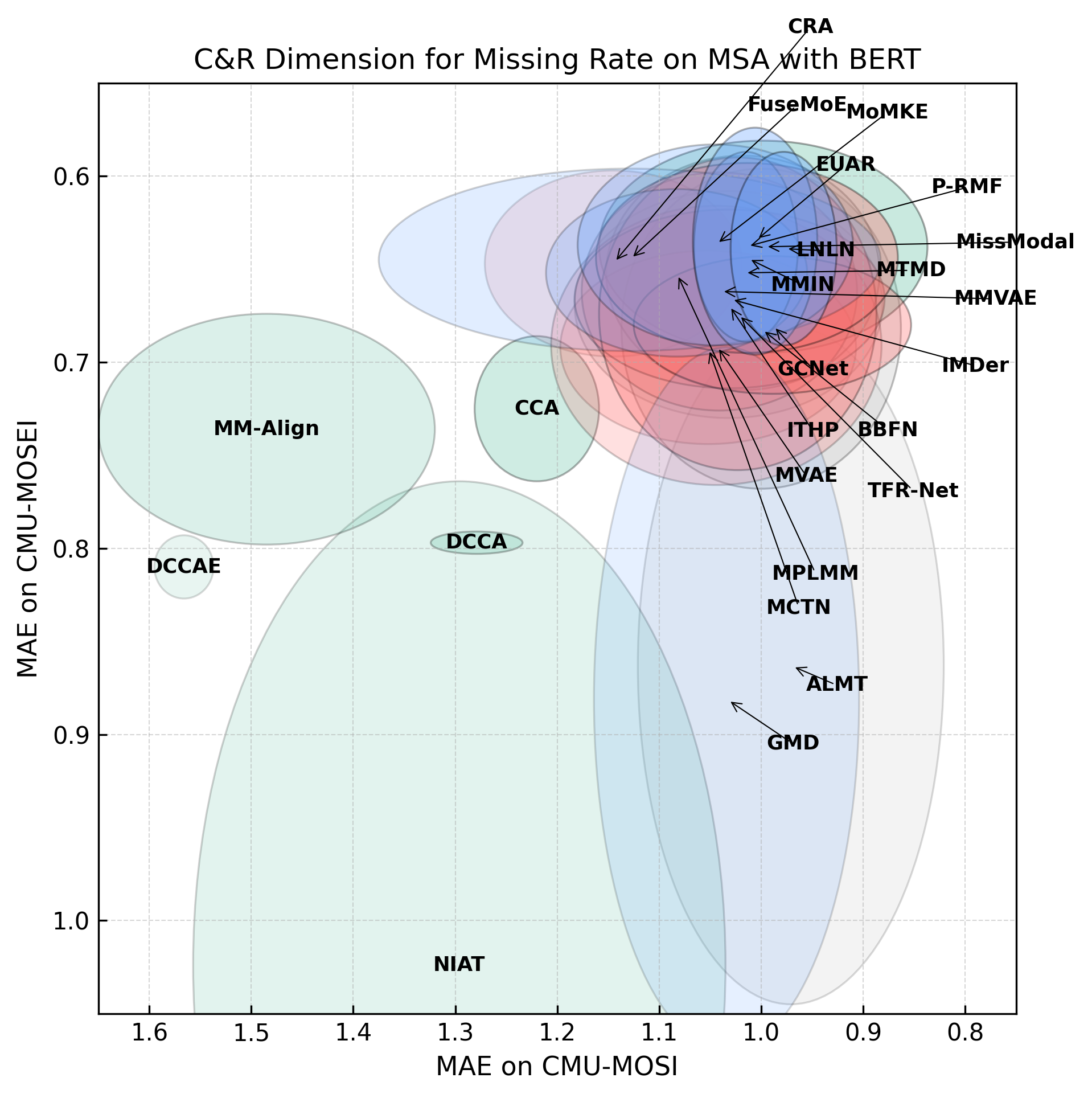}
}
\subfigure[with $MR$ on MER]{
\label{fig_vis_CR_dimension_BERT_MER_MR}
\includegraphics[width=0.23\textwidth]{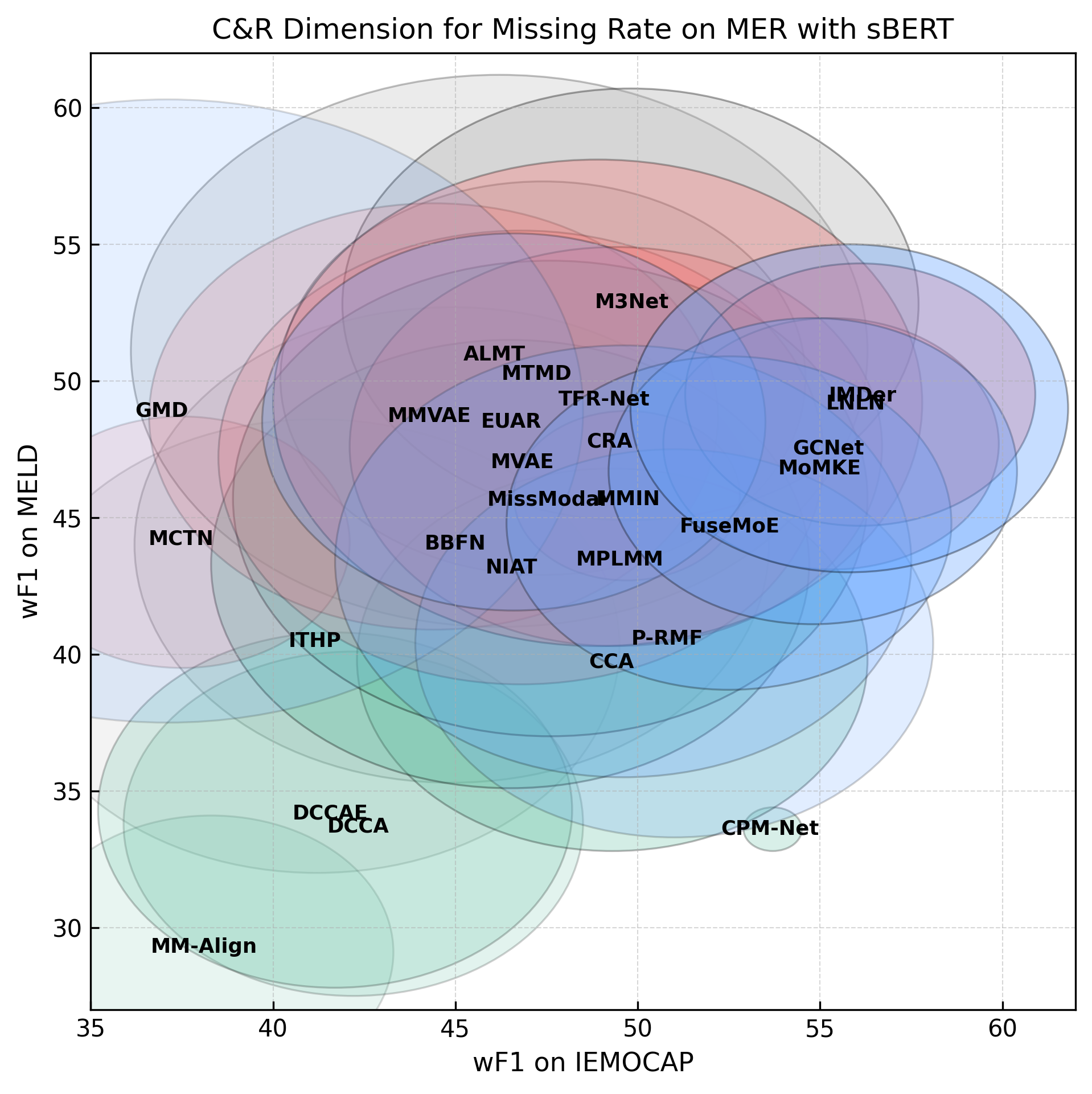}
}
\subfigure[with $MP$ on MSA]{
\label{fig_vis_CR_dimension_BERT_MSA_MP}
\includegraphics[width=0.23\textwidth]{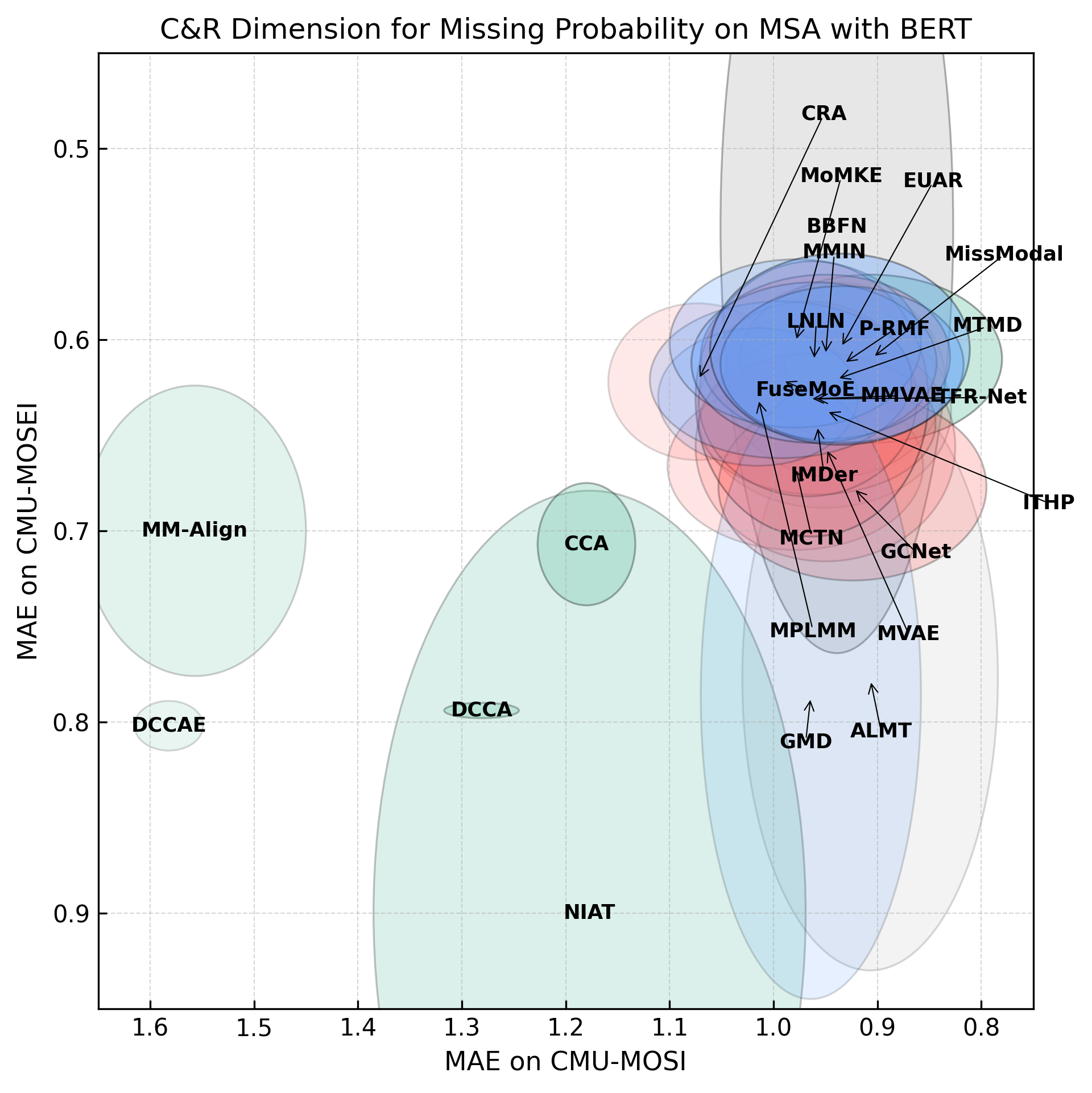}
}
\subfigure[with $MP$ on MER]{
\label{fig_vis_CR_dimension_BERT_MER_MP}
\includegraphics[width=0.23\textwidth]{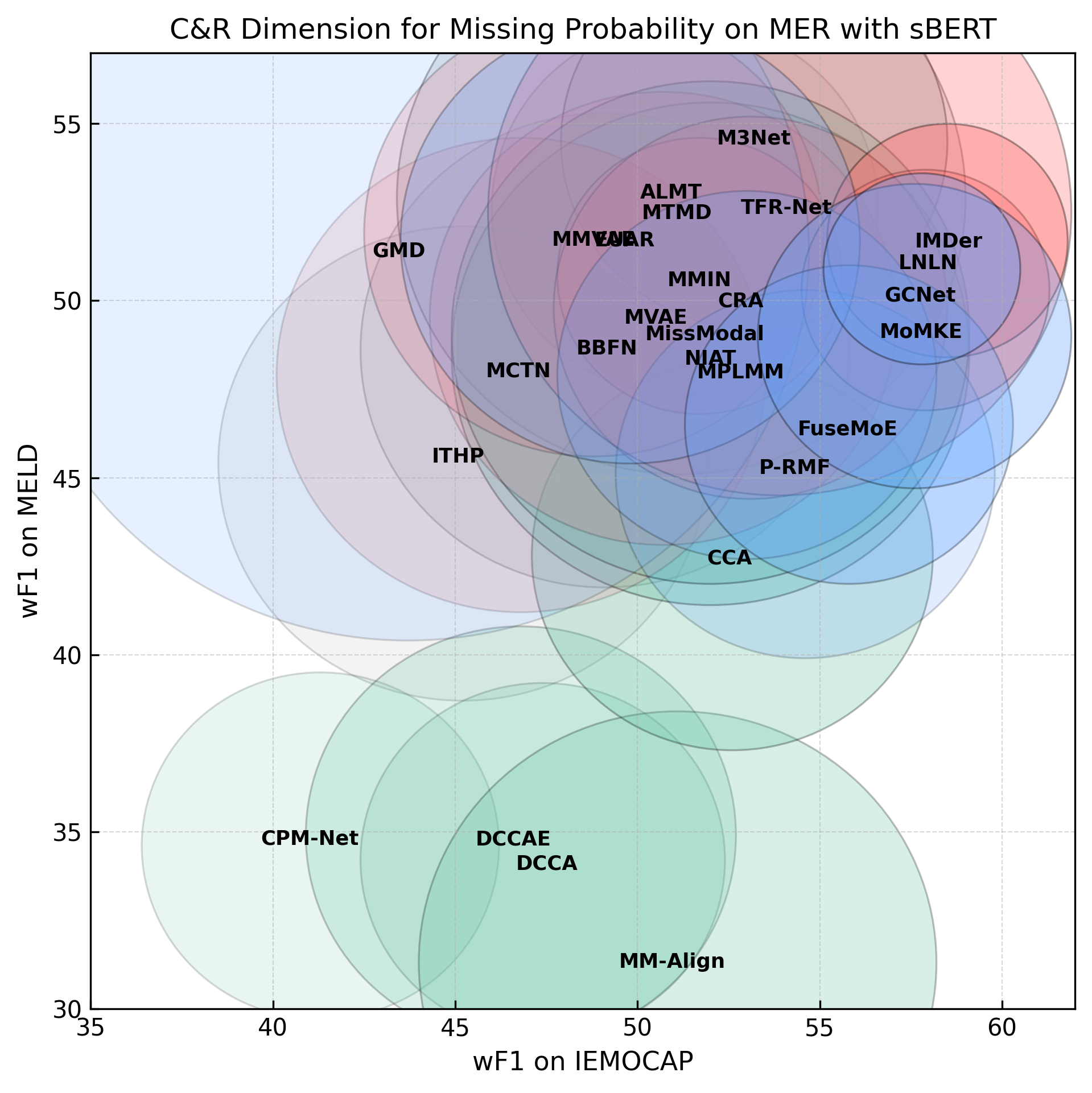}
}
\caption{Evaluation on $C\&R$ Dimension with BERT/sBERT on MSA and MER subtasks at both dataset-level ($MR$) and instance-level ($MP$) random missing protocols.}
\label{figure_vis_CR_dimension_BERT}
\end{figure*}

\subsection{Performance with $C\&R$ Dimension:}
The overall performance of each model across both dataset-level and instance-level random missing protocols is presented in Figure \ref{figure_vis_CR_dimension_BERT} with diverse colors denoting four types of models. We observe that adaptation-based methods and generation-based methods consistently cluster in the upper-right region with smaller circle sizes and concentrated positions across all scenarios, indicating they achieve both strong average performance and minimal fluctuation across diverse missing scenarios. While data augmentation based methods and alignment-based methods are dispersed across wider regions with larger axes ranges, which reflects their unstable performance in addressing dynamic missing scenario in the real world.

Besides, models exhibit distinct behavioral patterns between MSA and MER tasks. In sentiment regression of MSA, the performance relies more on the types of model with wider spread, while in emotion classification of MER, models cluster more tightly with generally higher and more similar fluctuation degree. MER demonstrates better overall robustness across diverse model types, with even alignment methods showing reduced fluctuation compared to MSA. This indicates that classification tasks are inherently more resilient to missing modalities than regression tasks where precise sentiment prediction requires more complete multimodal information. The tighter clustering in MER indicates that categorical multimodal emotion features remain more identifiable even with incomplete inputs, whereas continuous sentiment scoring in MSA suffers more dramatically from missing modality issue. 

Additionally, models show different stability patterns across MSA datasets, most superior methods exhibit relatively stable performance on MOSEI (smaller vertical axis) but larger fluctuation on MOSI (larger horizontal axis), as evidenced by the ellipses being more horizontally elongated. This suggests that incomplete multimodal learning can benefit from larger-scale dataset with more diverse missing samples to promote generalization ability. While for MER, IEMOCAP's more balanced emotion distribution leads to higher performance fluctuation (larger vertical axis) since models cannot heavily focus on specific dominant emotion classes to improve performance, requiring more robust multimodal understanding across all categories. Conversely, MELD's imbalanced distribution (smaller horizontal axis) allows models to achieve more stable performance by focusing on frequent emotion classes, resulting in reduced fluctuation but potentially less generalization. Thus, more empirical experiments and in-depth analyses are required in the future to address the missing modality issue in multimodal affective computing.



\section{Conclusion}
In this paper, we present MissMAC-Bench, a comprehensive benchmark for evaluating the missing modality issue in multimodal affective computing. By providing unified evaluation standards to simulate the real world scenarios, we integrate fixed and random missing protocols at dataset and instance levels, which evaluates both competence and resilience of MAC models in complete and incomplete scenarios. Extensive experiments are conducted on 4 types of models across 2 sub-tasks, 3 widely-used language models, and 4 targeted MAC datasets. Our work enables systematic evaluation of inter-modal robustness and cross-modal synergy, offering a solid foundation for developing more generalizable and robust MAC models.

\section*{Limitations}
Despite its comprehensive design, MissMAC-Bench has several limitations as the future improvement direction. First, while it focuses on modality-level missing scenarios that align with many real-world settings, future extensions could incorporate finer-grained missing patterns such as token-level missing, correlated missing patterns, or noisy input to enable more detailed analysis of intra- and inter-modal dynamics. Second, current benchmark centers on MSA tasks, while expanding the proposed standard to broader multimodal applications could enhance its generalizability. Third, although the benchmark has evaluated several representative encoder-based models in MSA, future work should include more advanced architectures, particularly decoder-based multimodal large language models, to better understand their robustness under missing modality conditions. 


\bibliography{missingmodality}

\appendix

\section{Further Discussion with Missing Degree}
\label{missing_degree_appendix}
\begin{figure}[htbp]
	\centering 
	\includegraphics[scale=0.55]{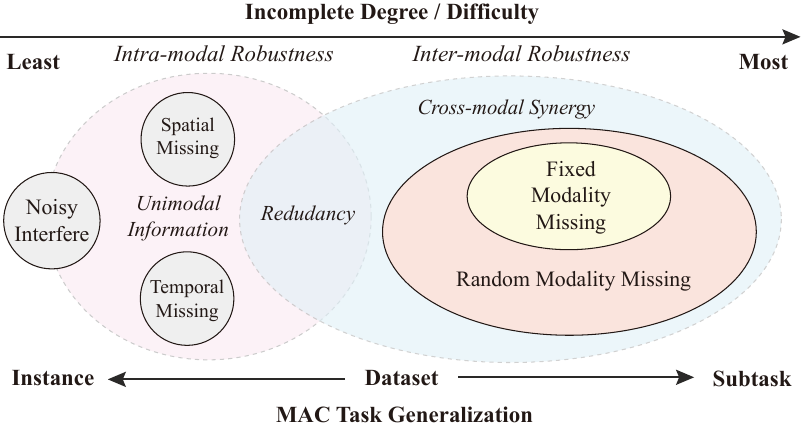}
	\caption{Incomplete multimodal input with diverse missing degrees and task generalization ability.}
	\label{figure_problem_definition}
\end{figure}

Based on the missing degree of modality data, we categorize incomplete multimodal input into intra- and inter-modal missing circumstances, as illustrated in Figure \ref{figure_problem_definition}. The intra-modal missing scenario evaluates the intra-modal robustness to either noisy inputs or spatial and temporal missing scenarios \cite{yuan2023noise}, where the unimodal contextual information is partially absent. In contrast, the inter-modal missing scenario simulates cases in which an entire modality is removed from the multimodal input. Intuitively, the inter-modal missing  presents greater difficulty, as it requires leveraging cross-modal interactions and compensating for the absence of entire modalities. Besides, inter-modal missing emphasizes the effect of cross-modal synergy, which plays a more complex and crucial role in multimodal fusion \cite{liang2023multimodal,fei2025on} from the perspective of information decomposition \cite{williams2010nonnegative,liang2023quantifying}. Thus in this paper, we focus on scenarios with whole modality missing to reveal model's inter-modal robustness and ability in cross-modal semantic understanding.

Beyond the simulated missingness settings in MissMAC-Bench, we plan to investigate more realistic and temporally structured modality degradation in future work, which further bridge the gap between the proposed missing simulation and the real world scenarios. Specifically, we will introduce modality-specific noise simulators, including misspelling, word-order perturbation, and character insertion for language, muting and background noise for audio, and black-screen, salt-and-pepper noise, Gaussian noise, and blur for visual inputs. We will further consider both fixed noise settings and probability or rate guided instance-level random noise, similar as the evaluation protocols in MissMAC-Bench. This extension will provide a more comprehensive evaluation of multimodal robustness under practical and heterogeneous failure conditions.

\section{Implementation Details}
All experiments are conducted on H800 GPU with CUDA 11.8 and Pytorch 2.0.1. Following \citep{gkoumas2021makes}, we perform fifty-times of random grid search for the best hyper-parameters of each model and report the average results over 10 seeds as the final performance.

\section{Dataset Details}
The details about the dataset distribution is introduced as Table \ref{dataset}. We cover MAC task with both MSA and MER subtasks on four datasets. Each subtask contains two size of dataset to reveal partial scalability performance of each model, such as MOSI and MOSEI in MSA. Besides, we consider the impact of class balance in classification for MER subtask, where IEMOCAP dataset has a more balanced  distribution on emotion classes while MELD dataset is collected with higher imbalance distribution.

\begin{table}[htbp]
\caption{Dataset distribution of MissMAC-Bench.}
\label{dataset}
\centering
\scalebox{0.8}{
\begin{tabular}{ccccc}
\toprule[1pt]
  MAC Task & \multicolumn{2}{c}{Sentiment Analysis} & \multicolumn{2}{c}{Emotion Recognition} \\ 
\cmidrule(lr){2-3}\cmidrule(lr){4-5}
  Dataset & MOSI & MOSEI & IEMOCAP & MELD \\ 
\midrule[1pt]
  Train & 1,284 & 16,326 & 5,228 & 9,765 \\
  Valid & 229 & 1,871 & 519 & 1,102 \\
  Test & 686 & 4,659 & 1,622 & 2,524 \\
\bottomrule[1pt]
\end{tabular}
}
\vspace{-0.2cm}
\end{table}

\section{More Experiment Results}
\label{more_exp_appendix}
We have released the evaluation code and all experiment results with different missing circumstances specifically in \url{https://github.com/RH-Lin/MissMAC-Bench}.

\begin{table*}[htbp]
\centering
\setlength\tabcolsep{4pt}
\caption{Performance comparison with complete input modalities set as $u\in\{l,a,v\}$. }
\label{table_exp_complete}
\scalebox{0.5}{
    \begin{tabular}{c|c|cccccccccccc|cccccccccccc}
    \toprule[1.5pt]
    \multicolumn{2}{c}{Language Model} & \multicolumn{12}{c}{BERT-base / sBERT-base} & \multicolumn{12}{c}{DeBERTaV3-large}\\
    \midrule[1.5pt]
    \multirow{2}{*}{Comp} & \multirow{2}{*}{Models} & \multicolumn{4}{c}{CMU-MOSI} & \multicolumn{4}{c}{CMU-MOSEI} & \multicolumn{2}{c}{IEMOCAP} & \multicolumn{2}{c}{MELD} & \multicolumn{4}{c}{CMU-MOSI} & \multicolumn{4}{c}{CMU-MOSEI} & \multicolumn{2}{c}{IEMOCAP} & \multicolumn{2}{c}{MELD}\\
    \cmidrule(r){3-6}\cmidrule(r){7-10}\cmidrule(r){11-12}\cmidrule(r){13-14}\cmidrule(r){15-18}\cmidrule(r){19-22}\cmidrule(r){23-24}\cmidrule(r){25-26}
    & & Acc7$\uparrow$ & F1$\uparrow$ & MAE$\downarrow$ & Corr$\uparrow$ & Acc7$\uparrow$ & F1$\uparrow$ & MAE$\downarrow$ & Corr$\uparrow$ & Acc$\uparrow$ & wF1$\uparrow$ & Acc$\uparrow$ & wF1$\uparrow$& Acc7$\uparrow$ & F1$\uparrow$ & MAE$\downarrow$ & Corr$\uparrow$ & Acc7$\uparrow$ & F1$\uparrow$ & MAE$\downarrow$ & Corr$\uparrow$ & Acc$\uparrow$ & wF1$\uparrow$ & Acc$\uparrow$ & wF1$\uparrow$\\

    \midrule[1.5pt]
     \multirow{27}{*}{$\{l,a,v\}$} 
     
    & BBFN$^\tau$ & 45.9 & 84.5 & 0.750 & 0.790 & 53.9 & 86.0 & 0.522 & 0.788 & 60.2 & 60.2 & 59.6 & 58.1 & 
    - & - & - & - & - & - & - & - & - & - & - & - \\ 
    & MTMD$^\tau$ & 47.5 & 85.3 & 0.705 & 0.799 & 53.7 & 86.0 & 0.535 & 0.760 & 60.1 & 59.9 & 57.7 & 58.0 & 
    - & - & - & - & - & - & - & - & - & - & - & - \\
    & ALMT$^\tau$ & 49.0 & 84.8 & 0.694 & 0.806 & 54.3 & 84.0 & 0.519 & 0.779 & 63.6 & 63.5 & 60.7 & 59.5 & 
    51.5 & 89.3 & 0.559 & 0.884 & 55.0 & 86.2 & 0.494 & 0.807 & 67.7 & 67.7 & 59.1 & 58.0\\
    & ITHP$^\tau$ & 44.5 & 83.6 & 0.758 & 0.779 & 52.0 & 84.7 & 0.551 & 0.756 & 56.5 & 56.2 & 56.6 & 56.6 &
    50.2 & 88.5 & 0.576 & 0.872 & 54.2 & 87.5 & 0.501 & 0.810 & 66.3 & 66.2 & 60.2 & 59.5\\
    & M3Net$^\tau$ & - & - & - & - & - & - & - & - & 59.9 & 60.3 & 63.4 & 61.2 & 
    - & - & - & - & - & - & - & - & - & - & - & - \\
    
    \cmidrule(r){2-26}

    & CCA & 27.7 & 74.9 & 1.106 & 0.541 & 46.1 & 82.9 & 0.654 & 0.666 & 61.7 & 61.5 & 51.2 & 46.6 & 
    24.2 & 74.9 & 1.167 & 0.492 & 44.0 & 77.3 & 0.701 & 0.565 & 58.1 & 57.2 & 50.5 & 45.5\\
     & DCCA & 25.1 & 73.6 & 1.220 & 0.422 & 39.3 & 73.3 & 0.787 & 0.425 & 56.8 & 55.5 & 47.7 & 37.0 &
     16.6 & 50.4 & 1.396 & 0.178 & 40.8 & 73.0 & 0.778 & 0.406 & 42.7 & 40.6 & 48.1 & 31.3\\
     & DCCAE & 19.7 & 69.5 & 1.642 & 0.357 & 38.9 & 73.5 & 0.782 & 0.437 & 57.4 & 56.4 & 48.0 & 36.9 &
     18.1 & 49.5 & 1.405 & 0.186 & 42.2 & 68.8 & 0.786 & 0.386 & 37.4 & 35.6 & 48.9 & 34.2\\
     & CPM-Net & 16.4 & 65.5 & 1.337 & 0.348 & 35.9 & 75.4 & 0.873 & 0.375 & 55.7 & 56.2 & 42.3 & 38.0 &
     15.5 & 50.7 & 1.405 & 0.092 & 41.4 & 48.5 & 0.839 & 0.132 & 26.3 & 18.2 & 28.2 & 26.8\\
     & MM-Align & 33.5 & 83.2 & 0.881 & 0.717 & 50.9 & 82.0 & 0.570 & 0.737 & 49.0 & 49.4 & 40.5 & 36.6 & 
    - & - & - & - & - & - & - & - & - & - & - & - \\
     & NIAT & 38.8 & 82.8 & 0.819 & 0.746 & 54.0 & 85.5 & 0.529 & 0.769 & 62.8 & 62.7 & 59.4 & 60.7 &
     50.9 & 90.2 & 0.565 & 0.883 & 47.0 & 75.1 & 0.679 & 0.567 & 62.5 & 62.6 & 58.3 & 58.4\\
     & MissModal & 47.1 & 86.0 & 0.698 & 0.801 & 53.9 & 85.8 & 0.533 & 0.769 & 63.5 & 63.7 & 57.1 & 57.0 & 
    - & - & - & - & - & - & - & - & - & - & - & - \\
    
    \cmidrule(r){2-26}

     & CRA & 34.8 & 83.2 & 0.916 & 0.741 & 51.4 & 85.5 & 0.553 & 0.765 & 63.4 & 62.2 & 57.6 & 54.8 & 
     46.8 & 87.2 & 0.615 & 0.861 & 55.6 & 88.3 & 0.498 & 0.804 & 60.8 & 58.7 & 62.1 & 60.6\\ 
     & MVAE & 43.7 & 84.3 & 0.745 & 0.780 & 52.9 & 84.4 & 0.554 & 0.772 & 60.9 & 61.1 & 54.2 & 54.5 & 
     - & - & - & - & - & - & - & - & - & - & - & - \\
     & MMVAE & 39.7 & 84.2 & 0.795 & 0.782 & 52.2 & 85.4 & 0.545 & 0.770 & 60.2 & 59.0 & 56.6 & 57.4 & 
     - & - & - & - & - & - & - & - & - & - & - & - \\
     & MCTN & 43.0 & 84.6 & 0.752 & 0.783 & 47.9 & 84.2 & 0.592 & 0.721 & 58.4 & 57.8 & 56.3 & 52.4 & 
     48.0 & 87.9 & 0.609 & 0.859 & 53.4 & 86.9 & 0.514 & 0.798 & 60.6 & 59.2 & 34.2 & 29.9\\
     & MMIN & 43.2 & 85.0 & 0.744 & 0.782 & 52.9 & 84.9 & 0.537 & 0.769 & 62.5 & 62.7 & 60.6 & 56.2 & 
     45.3 & 89.5 & 0.636 & 0.862 & 52.5 & 86.3 & 0.532 & 0.790 & 59.8 & 59.6 & 30.2 & 30.8\\
     & TFR-Net & 43.7 & 86.3 & 0.774 & 0.766 & 55.3 & 85.8 & 0.511 & 0.783 & 66.8 & 66.4 & 59.5 & 60.8 & 
     48.7 & 88.6 & 0.608 & 0.864 & 57.7 & 87.6 & 0.474 & 0.820 & 64.6 & 63.0 & 45.3 & 42.9\\
     & GCNet & 43.6 & 85.8 & 0.732 & 0.792 & 52.6 & 85.9 & 0.531 & 0.778 & 63.0 & 63.0 & 60.4 & 58.5 & 
     - & - & - & - & - & - & - & - & - & - & - & - \\ 
     & IMDer & 43.8 & 85.7 & 0.724 & 0.796 & 52.8 & 85.1 & 0.532 & 0.756 & 63.4 & 63.8 & 61.1 & 59.7 & 
     - & - & - & - & - & - & - & - & - & - & - & - \\

    \cmidrule(r){2-26}
    
     & MPLMM & 37.3 & 84.2 & 0.846 & 0.785 & 49.8 & 84.6 & 0.567 & 0.761 & 62.6 & 62.1 & 57.1 & 57.5 & 
     - & - & - & - & - & - & - & - & - & - & - & - \\
     & MoMKE & 44.5 & 80.3 & 0.784 & 0.758 & 53.1 & 85.9 & 0.528 & 0.776 & 64.6 & 64.6 & 57.1 & 56.3 & 
     - & - & - & - & - & - & - & - & - & - & - & - \\
     & FuseMoE & 45.0 & 85.2 & 0.736 & 0.789 & 52.0 & 85.0 & 0.553 & 0.747 & 63.7 & 63.2 & 47.1 & 49.5 & 
     - & - & - & - & - & - & - & - & - & - & - & - \\
     & GMD & 40.1 & 85.4 & 0.790 & 0.784 & 54.1 & 86.5 & 0.516 & 0.784 & 61.3 & 61.1 & 57.1 & 56.3 & 48.0 & 88.6 & 0.612 & 0.864 & 55.2 & 87.9 & 0.484 & 0.817 & 63.0 & 62.6 & 53.5 & 53.4 \\
     & EUAR & 46.2 & 85.3 & 0.725 & 0.792 & 54.0 & 86.1 & 0.522 & 0.785 & 59.9 & 60.1 & 58.6 & 57.5 & 50.8 & 89.9 & 0.566 & 0.884 & 54.6 & 87.6 & 0.498 & 0.813 & 58.3 & 58.5 & 59.2 & 58.8 \\
     & LNLN & 44.0 & 84.3 & 0.762 & 0.766 & 52.6 & 85.1 & 0.542 & 0.772 & 62.9 & 62.5 & 58.2 & 57.1 & 48.0 & 88.7 & 0.601 & 0.877 & 54.4 & 87.5 & 0.507 & 0.810 & 64.2 & 64.0 & 54.7 & 56.2\\
     & P-RMF & 43.0 & 84.0 & 0.744 & 0.788 & 51.8 & 85.5 & 0.543 & 0.775 & 63.7 & 63.0 & 56.0 & 55.6 & 
     - & - & - & - & - & - & - & - & - & - & - & - \\
    
    \bottomrule[1.5pt]
    
\end{tabular}
}
\end{table*}

\begin{figure*}[htbp]
\centering
\subfigure[With $MR$ on MSA]{
\label{fig_vis_CR_dimension_DeBERTa_MSA_MR}
\includegraphics[width=0.23\textwidth]{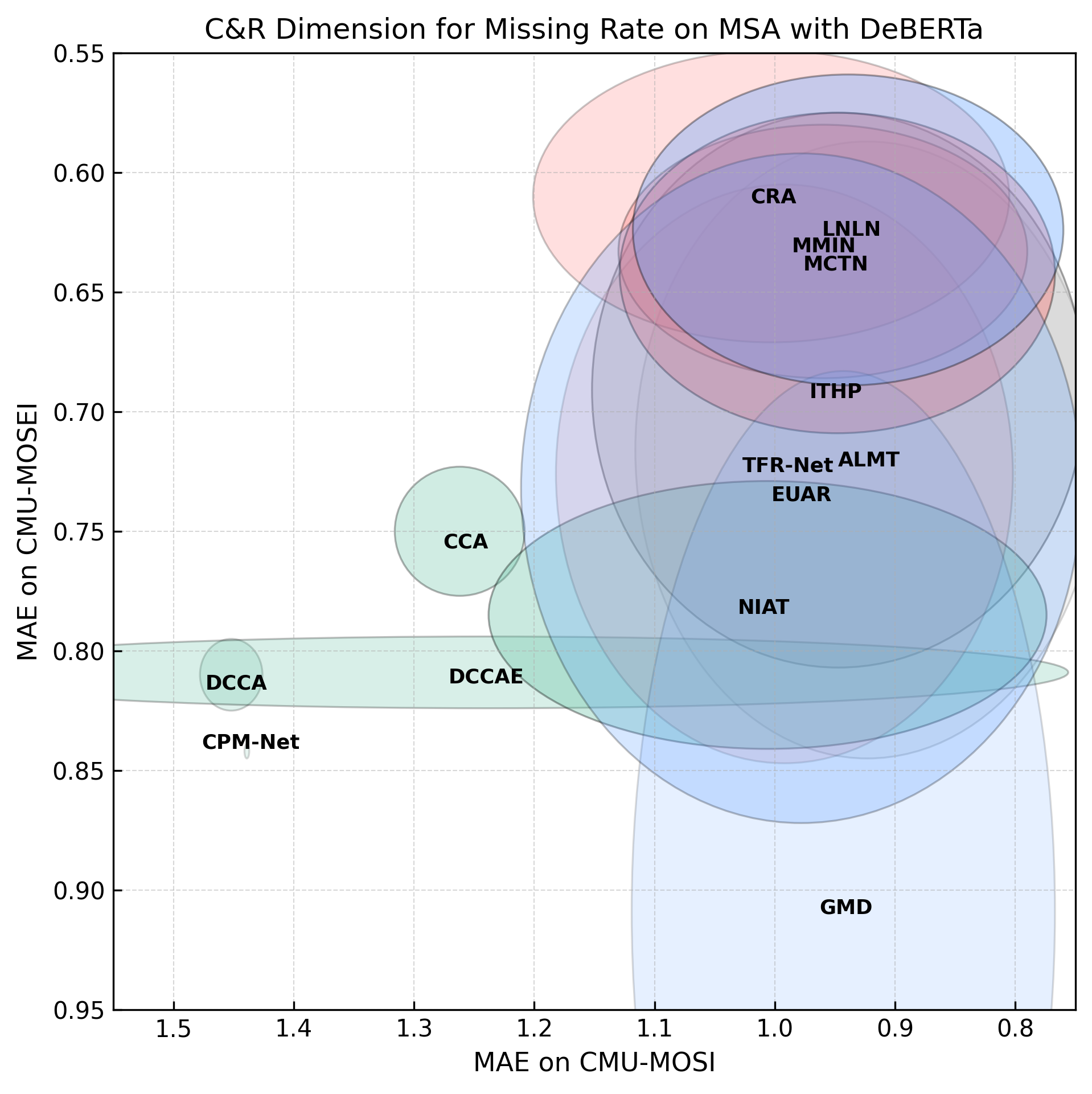}
}
\subfigure[With $MR$ on MER]{
\label{fig_vis_CR_dimension_DeBERTa_MER_MR}
\includegraphics[width=0.23\textwidth]{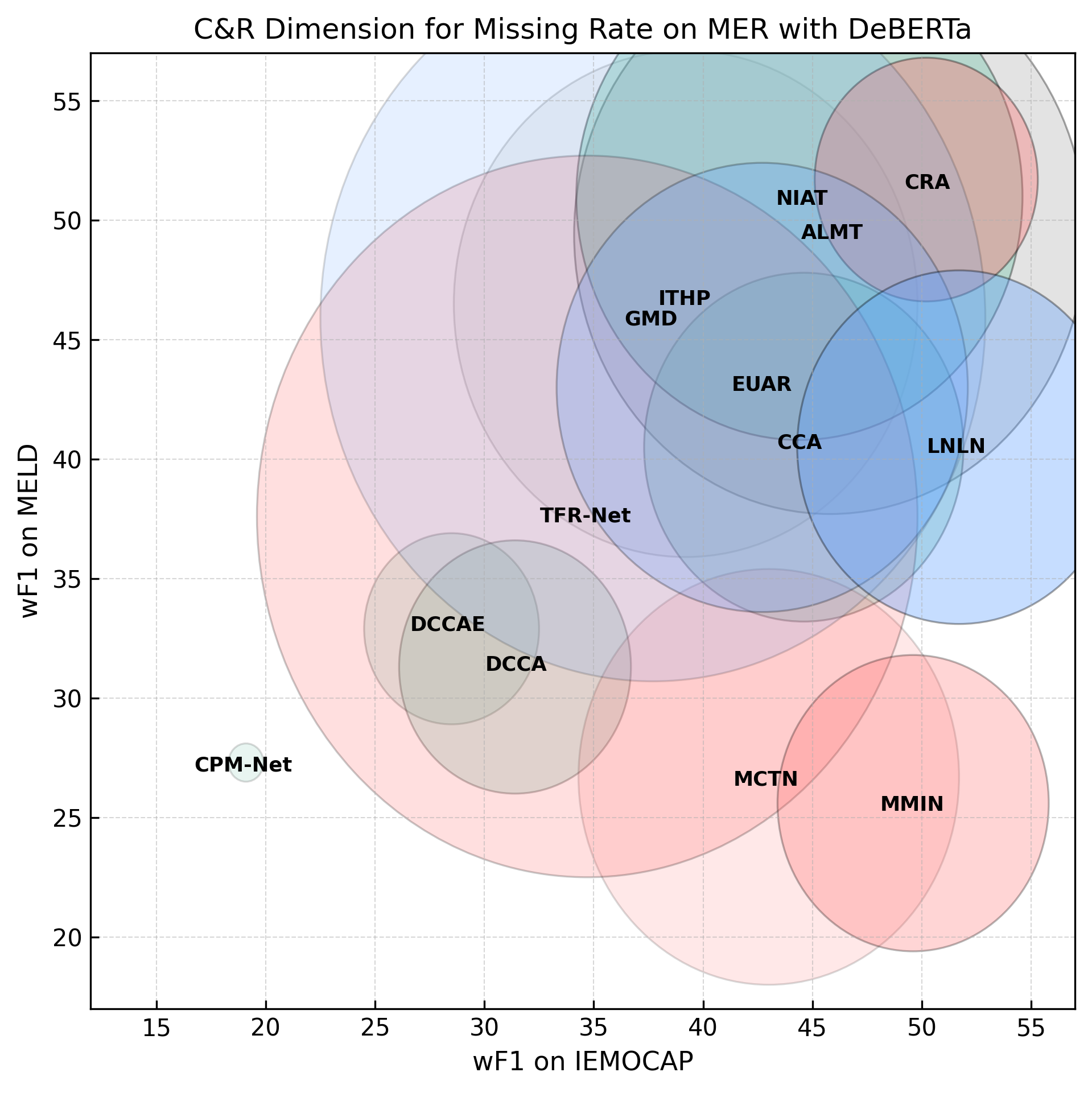}
}
\subfigure[With $MP$ on MSA]{
\label{fig_vis_CR_dimension_DeBERTa_MSA_MP}
\includegraphics[width=0.23\textwidth]{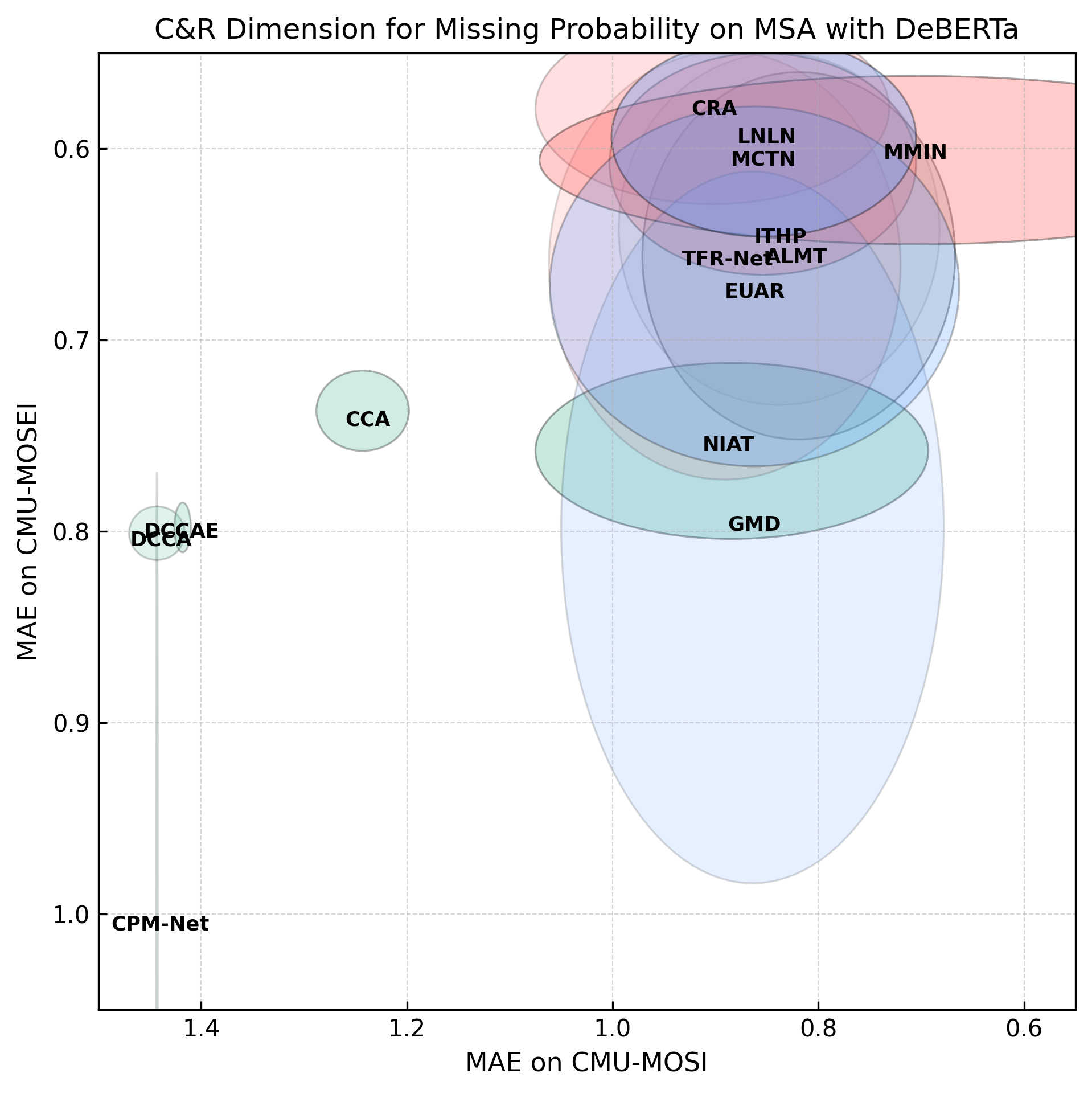}
}
\subfigure[With $MP$ on MER]{
\label{fig_vis_CR_dimension_DeBERTa_MER_MP}
\includegraphics[width=0.23\textwidth]{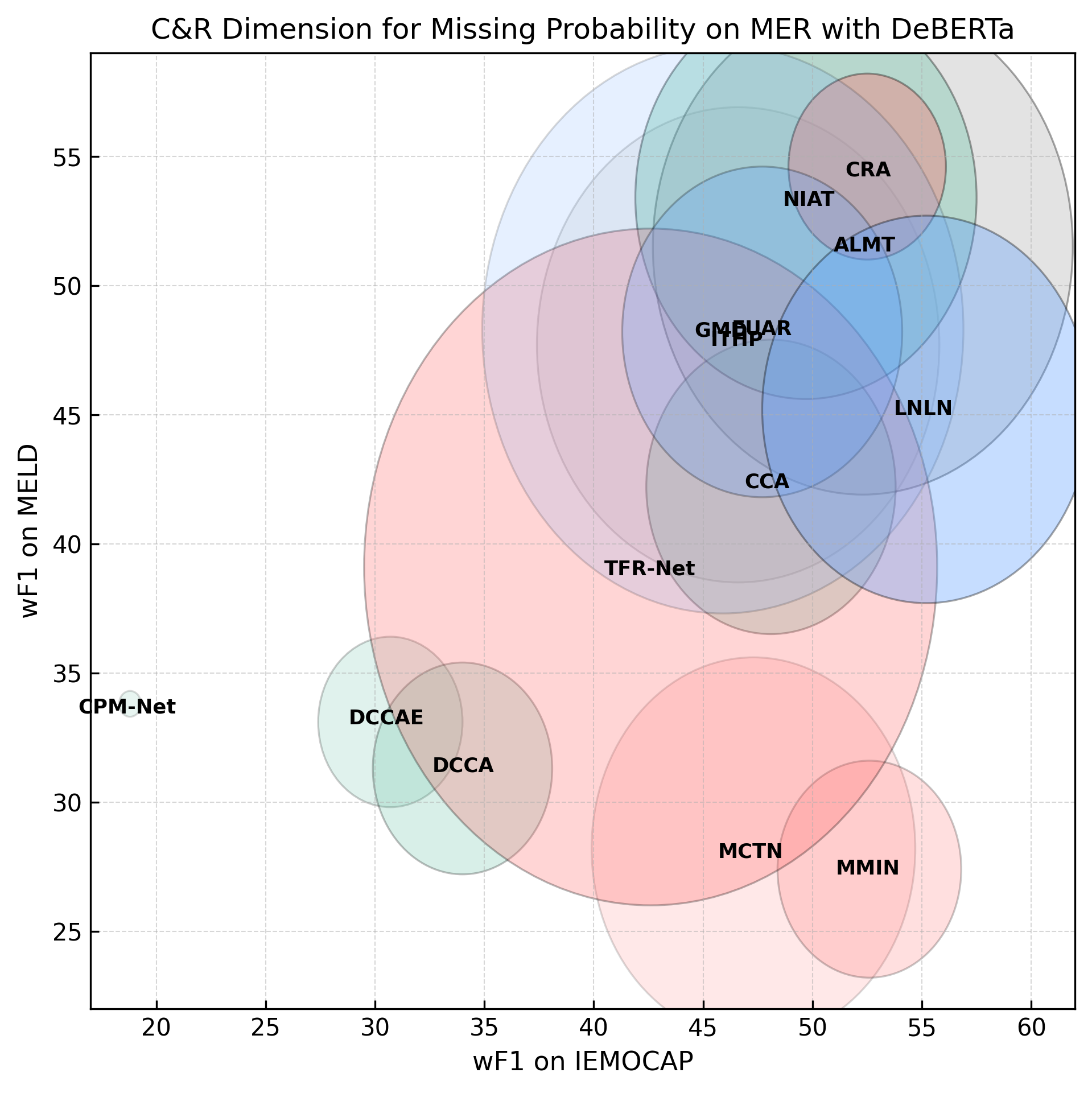}
}
\caption{Evaluation on $C\&R$ Dimension with DeBERTa on MSA and MER subtasks at both dataset-level ($MR$) and instance-level ($MP$) random missing protocols.}
\label{figure_vis_CR_dimension_DeBERTa}
\end{figure*}

\subsection{Performance with Complete Modalities}
We report experiment results with complete modalities as shown in Table \ref{table_exp_complete}. We can observe that performance of diverse models may vary significantly across different datasets. The state-of-the-art methods such as MTMD, ALMT and M3Net mostly reach the highest and balanced performance on both multimodal sentiment analysis and emotion recognition subtasks. 

\textbf{RQ1: Could larger language model achieve better performance?}  Larger language models demonstrate clear performance advantages in complete multimodal learning, as evidenced by DeBERTaV3 consistently outperforming BERT/sBERT across all metrics. For instance, ITHP achieves 88.5 F1 with DeBERTaV3 compared to 83.6 F1 with BERT on MOSI, representing approximately $5\%$ improvement. 

\textbf{RQ2: How does the data augmentation impact the state-of-the-art complete methods?} Compared with the performance reported in the original paper for the state-of-the-art methods such as ALMT or M3Net, leveraging data augmentation with missing modality mostly bring performance degradation on all metrics. We attribute it to the interference caused by the introduction of missing modal data on the fusion space of the original complete modalities.

\textbf{RQ3: How do different types of models perform when dealing with conflict between complete and incomplete learning?} Adaptation-based methods achieve the best balance on both MSA and MER subtasks due to the efficient fine-tuning techniques, while generative-based methods preserve the multimodal fusion weights which can also minimize interference of incomplete input. In contrast, alignment-based methods perform poorly overall in complete multimodal learning due to the large modality gaps, further highlighting the negative impact from incomplete multimodal learning to the complete one.

\subsection{$C\&R$ Dimension with DeBERTa}
We further provide the $C\&R$ Dimension with DeBERTa as shown in Figure \ref{figure_vis_CR_dimension_DeBERTa}. The trends across diverse datasets and evaluation levels remain largely consistent with those shown in Figure \ref{figure_vis_CR_dimension_BERT}. Nevertheless, we observe that larger language models tend to exhibit greater fluctuations and are highly dependent on the design of different model types, making it more challenging to enhance the robustness of multimodal models as the number of parameters increases. Therefore, research on addressing missing modality issue in the era of large language models still requires further investigation.







\end{document}